\documentclass[10pt,twocolumn,letterpaper]{article}

\usepackage[pagenumbers]{iccv} 

\usepackage[dvipsnames]{xcolor}

\newcommand{\systemName}{SLeRP\xspace}
\newcommand{\modelName}{Slot-Net\xspace}

\usepackage{pifont}
\usepackage{shortbold}
\newcommand{\cmark}{\textcolor{dark_green}{\ding{51}}}
\newcommand{\xmark}{\textcolor{red}{\ding{55}}}

\usepackage{xcolor}
\usepackage{colortbl}
\definecolor{red}{rgb}{0.8, 0.2, 0.2}
\definecolor{dark_green}{rgb}{0, 0.5, 0}

\usepackage{makecell}
\usepackage{adjustbox}
\usepackage{longtable}
\usepackage{graphicx}
\usepackage{afterpage}

\definecolor{rblue}{RGB}{68, 114, 196}
\definecolor{rgreen}{RGB}{112, 173, 71}
\definecolor{rorange}{RGB}{237, 125, 49}

\definecolor{iccvblue}{rgb}{0.21,0.49,0.74}
\usepackage[pagebackref,breaklinks,colorlinks,allcolors=iccvblue]{hyperref}
\usepackage{multirow}
\usepackage{float}
\usepackage{graphicx}
\usepackage{comment}

\makeatletter
\renewcommand\@makefnmark{}
\makeatother

\title{Slot-Level Robotic Placement via Visual Imitation from Single Human Video}

\author{Dandan Shan\textsuperscript{1,2\textsuperscript{*}} 
\quad
Kaichun Mo\textsuperscript{1\textsuperscript{\textdagger}}
\quad
Wei Yang\textsuperscript{1}\\
\quad
Yu-Wei Chao\textsuperscript{1}
\quad
David Fouhey\textsuperscript{3}
\quad
Dieter Fox\textsuperscript{1,4}
\quad
Arsalan Mousavian\textsuperscript{1}
\\
\hspace{0.11cm} \normalsize\textsuperscript{1} NVIDIA Research \quad    \textsuperscript{2} Univ.\ of Michigan \quad
\textsuperscript{3} NYU \quad
\textsuperscript{4} Univ.\ of Washington \quad
\\
\url{https://ddshan.github.io/slerp}
}

\begin{document}
\maketitle

\begin{abstract}
The majority of modern robot learning methods focus on learning a set of pre-defined tasks with limited or no generalization to new tasks. Extending the robot skillset to novel tasks involves gathering an extensive amount of training data for additional tasks. In this paper, we address the problem of teaching new tasks to robots using human demonstration videos for repetitive tasks (\eg, packing). This task requires understanding the human video to identify which object is being manipulated (the pick object) and where it is being placed (the placement slot). In addition, it needs to re-identify the pick object and the placement slots during inference along with the relative poses to enable robot execution of the task.
To tackle this, we propose \systemName, a modular system that leverages several advanced visual foundation models and a novel slot-level placement detector \modelName, eliminating the need for expensive video demonstrations for training. 
We evaluate our system using a new benchmark of real-world videos.
The evaluation results show that \systemName outperforms several baselines and can be deployed on a real robot.
\footnote{\textsuperscript{*} Work done during internship at Nvidia. \textsuperscript{\textdagger} Primary mentor.}
\end{abstract}

\vspace{-2mm}
\section{Introduction}
\label{sec:intro}
\vspace{-2mm}

Humans demonstrate exceptional skill in performing fine-grained manipulation tasks with high precision in their daily lives. 
From arranging eggs in an egg carton to sorting utensils in an organizer, we excel at tasks that require identifying and reasoning about which objects to pick up and how to place them into confined slots.
Cognitive and motor development theories suggest that we develop such skills at a young age, based on early experiences like playing with shape sorter toys~\cite{ornkloo2007fitting}.
However, current robotic and automated systems are not yet as adept as humans at perceiving and performing these fine-grained manipulation tasks.

\begin{figure}[!tbp]
\centering
\includegraphics[width=\linewidth]{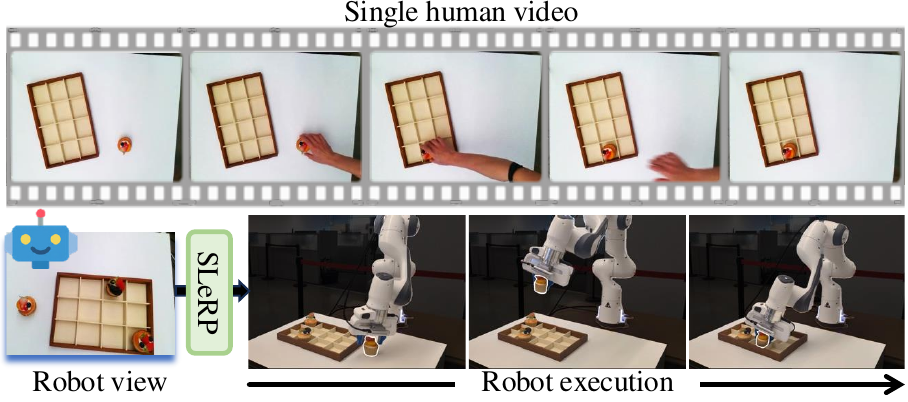}
\vspace{-7mm}
\caption{
We introduce the novel problem of imitating slot-level robotic placement from a single human video. 
Given a human demonstration video showing an object being placed in a slot, and a new robot-view image captured by the robot wrist camera (may feature varied camera and object poses, changed scenes), \systemName is able to find the corresponding object and similar slots in the robot view, and provide the 6-DoF transformation matrix for each detected slot to guide the robot in placing the object accurately.
}
\label{fig:teaser}
\vspace{-3mm}
\end{figure}

Slot-level manipulation is crucial in various industrial, logistics, and domestic contexts. 
For example, in industrial settings, machine tending~\cite{rooks2003machine} requires placing components precisely into machine slots for assembly or processing. 
In logistics, sorting and packaging tasks, such as organizing parcels in a warehouse or placing products into shipping containers, demand efficient and precise placement to optimize space and minimize damage. 
In domestic environments, future home assistant robots will need to perform slot-level manipulation tasks such as organizing items in cabinets, placing dishes in a dishwasher, and even preparing meals by accurately arranging ingredients in a pan.

The task of programming robots to perform slot-level placement remains arduous. 
Traditional methods~\cite{garrett:arcras2021,sundaralingam:icra2023} often require manual programming with domain expertise and assume that the object models and slot locations are known beforehand.
Learning-based approaches~\cite{brohan2022rt,kim2024openvla} show promise in alleviating the burden of programming; however, collecting robot data through tele-operation remains tedious and inefficient and can be particularly brittle for high-precision tasks due to embodiment gaps.
Learning from human demonstration videos has recently emerged as a promising approach due to its ease, speed of collection, and potential to capture slot-level details. 
However, previous research~\cite{xiong2021learning,bahl2022human,jain2024vid2robot,bharadhwaj2024towards,zhu2024vision} has generally been limited to coarser object-level tasks and often requires large amounts of training data to learn how to parse human demonstrations and translate them into robot policies. 

\begin{figure*}[!tbp]
    \includegraphics[width=\linewidth]{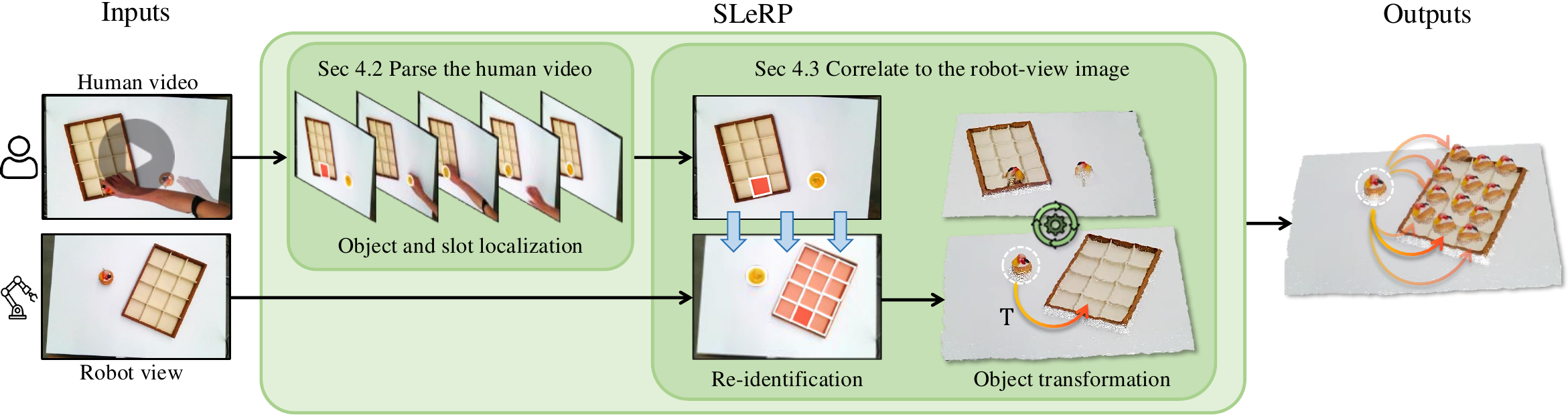}
    \vspace{-6mm}
    \caption{\textbf{Method Overview.} 
    The system begins by analyzing the input human video, tracking the object (highlighted in \textcolor[rgb]{1,0.69,0}{yellow}) throughout the sequence and identifying the placement slot (highlighted in \textcolor[rgb]{1,0,0}{red}). Next, we re-identify the object and the slot in the robot's view by correlating the human-view and robot-view images. Using depth images, we reconstruct the observations in 3D and compute a single 6-DoF object transformation $T$ in the robot's view, enabling the robot to transfer the object into the slot. If more than one slot is present, we detect all applicable slots and compute one 6-DoF object transformation for each slot. Finally, such 6-DoF object transformations are sent to the downstream robot planning and control pipeline for real robot pick-and-place execution.
    }
    \vspace{-3mm}
    \label{fig:overview}
\end{figure*}

In this paper, we study the novel problem of recognizing slot-level object placement from a \textit{single} human video, and estimating 6-DoF transformations for robot imitation. 
As shown in Fig.~\ref{fig:teaser}, the task takes two visual inputs: (1) a single human RGB-D video in which a person demonstrates picking up an object (\eg, a muffin) and precisely placing it into a slot within a placement object (\eg, a tray), and (2) a single RGB-D image captured from the robot's wrist camera, representing the new setup for the robot to operate in with possible varying camera and object poses compared to the human video.
The outputs aim to detect the object and all empty slots in the robot's view similar to the placement slot in the human video, as well as compute the 6-DoF object transformations necessary for the robot to transfer the object from its initial position to each of the slots.

We propose a novel modular approach called \systemName (\ie, \textbf{S}lot-\textbf{Le}vel \textbf{R}obotic \textbf{P}lacement), to tackle the problem.
As shown in Fig.~\ref{fig:overview}, \systemName starts by analyzing the input human demonstration video, tracking the manipulated object across the video frames and identifying the placement slot. 
Next, within the robot's view, \systemName re-identifies both the object and the slot by correlating the human-view images with the robot-view images.
By lifting the observations in 3D using the depth sensing and camera parameters, \systemName calculates a 6-DoF transformation matrix for the robot to transfer the object from its initial location to the desired slot in the robot's view.
If multiple slots are present, \systemName detects all slots that are similar to the one in the human video and computes the object transformations for all of them.
Finally, the computed 6-DoF object transformations are sent to the downstream robot planning and control pipeline for robot pick-and-place execution.

A key component of \systemName is the detection of placement slots. 
Currently, no existing method is specifically designed for this task, and simple image differencing or change detection~\cite{fang2022changer} does not effectively solve the problem.
Therefore, we propose a new slot-level placement detector, \modelName, that takes two image frames from a human demonstration video---one before and one after placement---and outputs a 2D mask outlining the placement slot on the images. Unlike common vision tasks, collecting a sizable training dataset for slot-level placement detection is challenging.
To address this challenge, we introduce a generative AI-based data creation pipeline that expands the 
training set by bootstrapping from a small set of images.

For evaluation, we introduce a new dataset comprising 288 real-world videos targeted at studying this novel problem.
We compare our method against several baseline approaches, including ORION~\cite{zhu2024vision}, a state-of-the-art method for object-level pick-and-place from a single human video; CLIPort~\cite{shridhar2021cliport}, an end-to-end imitation-learning-based language-conditioned policy for tabletop tasks; adapted versions of both for the novel slot-level task; and a custom baseline leveraging cutting-edge vision-language models like GPT-4o~\cite{hurst2024gpt}.
Our results demonstrate that \systemName outperforms baselines in accurately predicting placement slots and computing 6-DoF transformations across diverse real-world tasks.
Our ablation studies further validate several key components and design choices in our system. 
Finally, we conduct real-robot experiments that successfully apply the system in real world scenarios.

In summary, the core contributions of this paper are:
\begin{itemize}
\item Studying the novel task of slot-level object placement by learning from a single human demonstration video; 
\item Designing the modular approach \systemName and the slot-level placement detector \modelName to tackle this problem;
\item Introducing a new benchmark and several baseline methods to systematically evaluate system performance;
\item Demonstrating that \systemName achieves strong performance in real-world and real-robot evaluations.
\end{itemize}

\vspace{-1mm}
\section{Related Work}
\label{sec:related}
\vspace{-1mm}

\noindent
{\bf Object Placement in Robotics.}
Identifying where and how to place an object after picking it up is a crucial step in robotic pick-and-place tasks~\cite{lozano1989task}.
Early works~\cite{harada2014validating,baumgartl2014fast,haustein2019object} analytically search for flat features on the object and the placement surface.
Modern learning-based methods estimate placement locations and poses using learned features, focusing mainly on flat surfaces~\cite{newbury2021learning}, such as tabletop~\cite{mitash2020task,danielczuk2021object,yuanm2t2} and furniture shelves~\cite{murali2023cabinet}.
Researchers have also explored tabletop object placement under spatial and semantic constraints given other objects~\cite{paxtonpredicting,mees2020learning,liu2022structformer}.

In the more challenging case of placing an object on another non-flat object, prior work~\cite{jiang2012learning,sun2014object,simeonov2022neural,pan2023tax,simeonov2023se,eisner2024deep,zhu2024vision,sharma2024octo+,wang2024d} has explored tasks like putting one spoon in a cup or hanging a mug on a rack.
Our work extends these studies by focusing on placing objects into all empty, fine-grained, tight-fitting slots (\eg, all egg slots in a carton), a task that demands greater precision in recognition and prediction, as well as handling multiple placement locations.
Additionally, unlike previous work~\cite{zakka2020form2fit,zeng2021transporter,huang2022equivariant,wu2022transporters,shridhar2021cliport} that addresses a 2D planar setting and requires task-specific training from a few robot demonstrations, our approach tackles this problem in 3D, learning from a single human demonstration to enable one-shot generalization to novel tasks.

\vspace{1mm}
\noindent
{\bf Imitation Learning from Human Videos.}
Human videos serve as a natural, information-rich, and easily accessible source of data for learning robotic manipulation.
Previous work has explored diverse methods to extract, represent, and apply knowledge from human videos to support robot manipulation learning. 
These approaches include pre-training latent visual representations~\cite{nair2023r3m,dasari2023unbiased,majumdar2023we,shangtheia}, 
inferring action trajectories or plans~\cite{lee2017learning,nguyen2018translating,bahl2022human,wang2023mimicplay,ye2024latent}, 
learning value or reward functions~\cite{mavip,chen2021learning}, 
reconstructing human hand or hand-object interaction~\cite{qin2022dexmv,Patel2022,mandikal2022dexvip,shaw2023videodex,singh2024hand}, 
parsing interaction goals and affordance~\cite{liu2018imitation,yang2015robot,kannan2023deft,bahl2023affordances}, 
learning point tracks for human-to-robot transfer~\cite{xiong2021learning,wen2023any,bharadhwaj2024track2act,xuflow,yuan2024general}, etc.
While the primary goals of these works are typically learning robot trajectories or manipulation policies, our work explores a novel perspective by recognizing fine-grained placement slots as visual imitation targets.

Additionally, we tackle robot imitation learning from a single human video.
Previous work has investigated one-shot~\cite{yu2018one,dasari2021transformers,jang2022bc,mandi2022towards,jain2024vid2robot} and even zero-shot learning from human videos~\cite{bharadhwaj2024towards}; however, these approaches often require extensive human video datasets, sometimes paired with robot videos, to span multiple tasks during training.
In contrast, our approach leverages existing visual foundation models, eliminating the need for large-scale training videos.
A notably similar work ORION~\cite{zhu2024vision} relies on text to recognize task-relevant objects and primarily focuses on object-level pick-and-place.
In contrast, our method exclusively extracts information from a single human video to perform more fine-grained slot-level placement tasks.

\vspace{-2mm}
\section{Problem Formulation}
\label{sec:problem}
\vspace{-2mm}

We formulate the novel problem of recognizing slot-level object placement from a single human video, and estimating 6-DoF transformations for downstream robot imitation.

\vspace{1mm}
\noindent
{\bf Inputs.}
The task takes the following inputs:
\begin{itemize}
\item a single RGB-D human demonstration video with $n$ frames, denoted as $\mathbf{\mathcal{H}}=\{\mathbf{H}_1, \mathbf{H}_2, \cdots, \mathbf{H}_n\}$, recording a person picking up an object $O$ from the scene and placing it in a slot $S$ within a placement object;
\item a single RGB-D robot-view image $\mathbf{R}$ that captures the robot's observation, often taken from the robot wrist camera and possibly with different camera and object poses, or scene layouts.
\end{itemize}

\vspace{1mm}
\noindent
{\bf Outputs.}
The task outputs, in the robot's view, are:
\begin{itemize}
\item an object mask $\mathbf{M}^O_{\mathbf{R}}$ over the robot image $\mathbf{R}$ that segments the object $O$ to pick;
\item a list of slot masks $\{\mathbf{M}_R^{S_0},\mathbf{M}_R^{S_1}, \cdots, \mathbf{M}_R^{S_{k}}\}$ over the robot image $\mathbf{R}$ that marks all empty slots on the placement object similar to the demonstrated placement slot in the human video $\mathbf{\mathcal{H}}$;
\item a list of 3D 6-Degree-of-Freedom (DoF) object transformation matrices $\{\mathbf{T}_0,\mathbf{T}_1,\cdots,\mathbf{T}_k \mid \mathbf{T}_i \in SE(3)\}$ in the robot's coordinate frame, for the robot to transfer the object $O$ from its initial position to all the detected slots.
\end{itemize}
Passing the detected object and slot masks, as well as the calculated 6-DoF object transformatrion matrices, downstream robot pick-and-place pipeline is able to execute slot-level object placement as shown in Fig.~\ref{fig:teaser}.
\vspace{-2mm}
\section{Method}
\label{sec:method}
\vspace{-2mm}
In this section, we present the technical designs of \systemName.
We begin with an overview (Sec.~\ref{subsec:System_Overview}) and then dive into more details in parsing the input human video (Sec.~\ref{subsec:parsing_human}) and correlating to the robot's view image (Sec.~\ref{subsec:correlate_robot}).

\vspace{-1mm}
\subsection{System Overview}
\label{subsec:System_Overview}
\vspace{-1mm}

Taking as inputs a human demonstration video $\mathbf{\mathcal{H}}$ and a robot-view image $\mathbf{R}$,
our method \systemName~(Fig.~\ref{fig:overview}) 
starts with parsing the input human video (Sec.~\ref{subsec:parsing_human}) by tracking the object $O$ throughout the video frames and identifying the placement slot $S$.
After this process, we obtain an object mask $\mathbf{M}^O_{\mathbf{H}_1}$ and a slot mask $\mathbf{M}^S_{\mathbf{H}_1}$ over the first frame of the human video $\mathbf{H}_1$.
Next, by leveraging this information, \systemName correlates the human-view and robot-view images (Sec.~\ref{subsec:correlate_robot}), and re-identify the object mask $\mathbf{M}_\mathbf{R}^O$ and the slot mask $\mathbf{M}_\mathbf{R}^{S_0}$ in the robot-view image $\mathbf{R}$, as observed in the first human frame.
If multiple similar slots are present, the system detects other empty slots $\{\mathbf{M}_R^{S_1}, \cdots, \mathbf{M}_R^{S_{k}}\}$ as well.
Then, the system lifts the human and robot observations in 3D using the depth sensing and camera intrinsics, and computes a single 6-DoF object transformation matrix $\mathbf{T}_i \in SE(3)$ for each detected slot $\mathbf{M}_R^{S_i}$.

\begin{figure}[!tbp]
    \includegraphics[width=\linewidth]{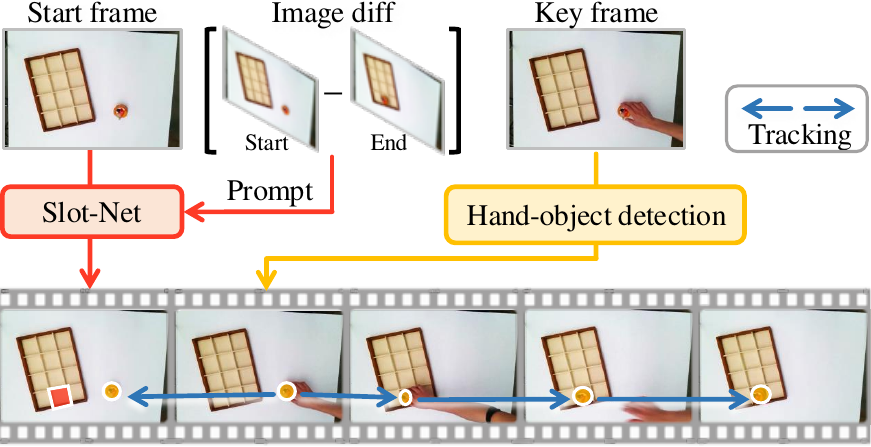}
    \vspace{-7mm}
    \caption{\textbf{Parse Human Video.} Given the input human video (bottom), we run state-of-the-art hand-object detector (\textcolor[rgb]{1,0.69,0}{yellow}) and tracker (\textcolor[rgb]{0,0,1}{blue}) to obtain the pick object mask (\textcolor[rgb]{1,0.69,0}{yellow}) and train a novel network \modelName (\textcolor[rgb]{1,0,0}{red}) to identify the slot mask (\textcolor[rgb]{1,0,0}{red}).
    }
    \label{fig:parse_human}
    \vspace{-4mm}
\end{figure}

\vspace{-1mm}
\subsection{Parsing the Human Demonstration Video}
\label{subsec:parsing_human}
\vspace{-1mm}
The input human video precisely demonstrates what is the pick object $O$ and where is the placement slot $S$.
As shown in Fig.~\ref{fig:parse_human}, our method utilizes powerful hand-object detection and tracking systems to identify the object mask $\mathbf{M}^O_{\mathbf{H}_1}$ and proposes a novel network \modelName for estimating the slot mask $\mathbf{M}^S_{\mathbf{H}_1}$ over the first frame of the human video.

\vspace{1mm}
\noindent
{\bf Object detection and tracking.} 
We use a hand-object detector~\cite{cheng2023towards} to detect frame-wise hands and in-contact objects, enabling us to locate the pick object $O$ in the human video.
As the detector operates on a per-frame basis, there may be temporally inconsistent predictions. To refine the detection results, we apply MASA's matching algorithm~\cite{li2024matching} to generate smooth trajectories for the hand and pick object across the hand-object contact frames. We then identify a confident key frame, when the hand and object first interacts, and use SAM2~\cite{ravi2024sam} to track through the video, producing per-frame object segmentation $\MB_{\HB_i}^O$.

\vspace{1mm}
\noindent
{\bf Placement slot detection (\modelName).}
Since no prior work has studied the problem of detecting the placement slot given a human pick-and-place video, we propose our own novel network \modelName for this purpose.
We leverage the SAM architecture~\cite{kirillov2023segment} as the backbone given its powerful capability in segmentation. 
\modelName takes the starting frame $\mathbf{H}_1$ of the pick-place video as the input, together with the absolute image difference in gray-scale between the starting and end frame $\left|\mathbf{H}_1-\mathbf{H}_n\right|$ as the visual prompt, and is tasked to output a slot segment in the starting human frame image $\mathbf{M}^S_{\mathbf{H}_1}$. We leverage SAM's large-scale pretraining by preserving most of its designs (\eg, the image encoder, the mask decoder) and we use the same image encoder as the prompt embedder to process image difference prompt. 
Since we find that the SAM pretraining does not directly work on such customized new task, finetuning over such slot-level placement data is necessary.

\begin{figure}[!tbp]
    \includegraphics[width=\linewidth]{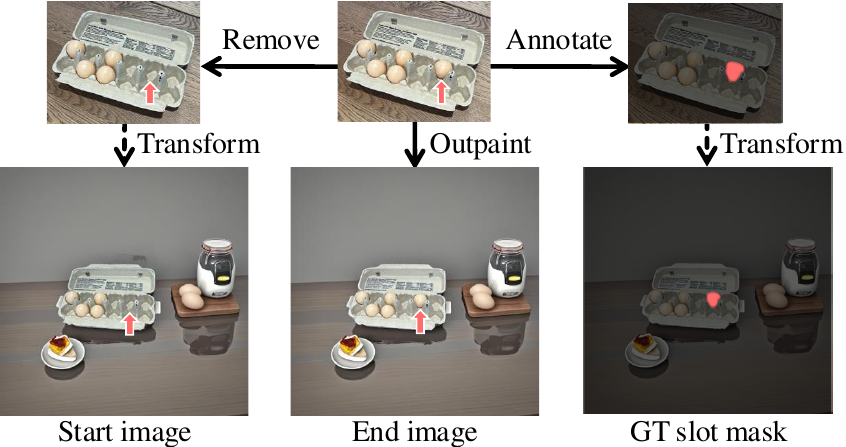}
    \vspace{-7mm}
    \caption{\textbf{\modelName Data Generation.} 
    Given an object-centric image (top middle), we inpaint to remove an object and reveal its slot (top left) and manually annotate the slot mask (top right). We then outpaint these images with a scene background (bottom) to create a starting and end image pair with a ground-truth slot mask.
    }
    \label{fig:data_gen}
    \vspace{-4mm}
\end{figure}

\vspace{1mm}
\noindent
{\bf \modelName data generation.}
Training our SAM-based \systemName requires a lot of data, yet collecting fine-grained slot-level placement data in the real world is expensive.
However, recent generative models have demonstrated great capabilities in generating realistic images~\cite{podellsdxl}, excelling at tasks such as object removal and image outpainting. 
We therefore propose a semi-automatic synthetic data generation pipeline (Fig.~\ref{fig:data_gen}).
Given a collected object-centric image of a placement object with many slots, we utilize a state-of-the-art object removal model (SDXL~\cite{von-platen-etal-2022-diffusers} and Cleanup.pictures~\cite{cleanup}) to remove one pick object from one slot and manually annotate the slot mask for the removed object using TORAS~\cite{toras}.
Then, we employ a powerful image outpainting generative model (Hugging Face Outpainting script~\cite{huggingface_outpainting}) to expand the image canvas, generating 100 images in diverse backgrounds, prompted with Llama~\cite{touvron2023llama} generated text prompts, for each object-centric image.

In this manner, we obtain a large number of annotated starting and end image pairs to train \systemName.
We crowdsourced and collected 2,138 object-centric images of items with slots, spanning 67 object categories, by capturing them in everyday environments.
We applied 100 augmentations for each slot on the object-centric image, resulting in 156K images for training, with the rest left for testing and validation.
See supplementary for more details.

\begin{figure*}[!t]
    \includegraphics[width=\linewidth]{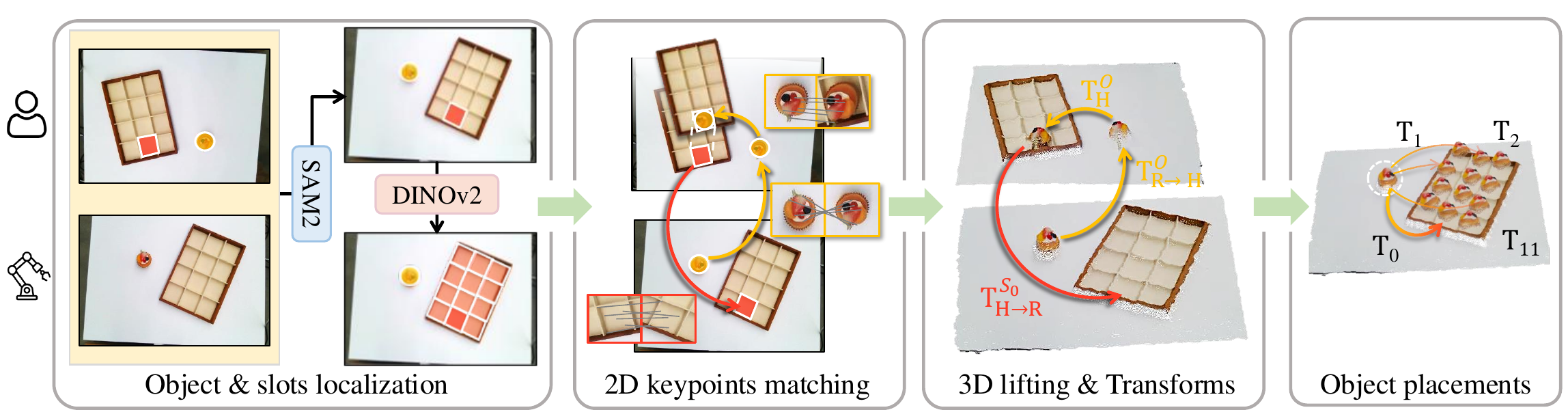}
    \vspace{-6mm}
    \caption{\textbf{Correlate with robot view.} 
    Given the object and slot mask detected in the human video, we first re-identify the corresponding object and slot in robot view, and also find all similar empty slots. With corresponding object masks and slot masks, we first compute 2D keypoint matching among the detected object and mask local patches and then lift the observations to 3D to compute 6-DoF transforms.
    }
    \label{fig:correlate_robot}
    \vspace{-3mm}
\end{figure*}

\subsection{Correlating to the Robot-view Image}
\label{subsec:correlate_robot}
\vspace{-1mm}
After we obtain the pick object mask $\mathbf{M}^O_{\mathbf{H}_1}$ and the slot mask $\mathbf{M}^S_{\mathbf{H}_1}$ from the human video, the next step is to correlate this information to the robot's view $\mathbf{R}$. 
As shown in Fig.~\ref{fig:correlate_robot}, \systemName first re-identifies the object mask $\mathbf{M}_\mathbf{R}^O$ and a list of empty slot masks $\{\mathbf{M}_\mathbf{R}^{S_0},\mathbf{M}_\mathbf{R}^{S_1}, \cdots, \mathbf{M}_\mathbf{R}^{S_k}\}$ similar to the human demonstrated placement slot.
Then, 2D keypoint matching and 3D lifting enable the calculation of a single 6-DoF object transformation matrix $\mathbf{T}_i \in SE(3)$ for each detected slot $\mathbf{M}_R^{S_i}$ for downstream robotic pick-and-place.

\vspace{1mm}
\noindent
{\bf Object and slot re-identification.}
Taking as input a short video with only two frames $\{\mathbf{H}_1, \mathbf{R}\}$, SAM2~\cite{ravi2024sam} is employed to output the object mask $\mathbf{M}_\mathbf{R}^O$ and one best-matched slot mask $\mathbf{M}_\mathbf{R}^{S_0}$ over the robot image $\mathbf{R}$, given the detected 2D object mask $\mathbf{M}^O_{\mathbf{H}_1}$ and the slot mask $\mathbf{M}^S_{\mathbf{H}_1}$ on the human first frame image $\mathbf{H}_1$.
If multiple similar slots are present in the robot image, we leverage SAM~\cite{kirillov2023segment} to propose segment candidates and use DINOv2~\cite{oquabdinov2} to collect additional slot masks $\{\mathbf{M}_\mathbf{R}^{S_1}, \cdots, \mathbf{M}_\mathbf{R}^{S_k}\}$ that share similar DINOv2 features with the detected slot mask $\mathbf{M}_\mathbf{R}^{S_0}$.
Empirically, we find that SAM2 and DINOv2 provide good enough performance on our data.

\vspace{1mm}
\noindent
{\bf 2D keypoint matching.}
With two corresponding masks in the human view and robot view, we use MASt3R~\cite{leroy2024grounding} to detect 2D keypoint correspondences by expanding the masks into local 2D bounding boxes.
As illustrated in Fig.~\ref{fig:correlate_robot} (middle left), we compute the 2D keypoint matching on two pairs of object local patches (between the object mask $\mathbf{M}_\mathbf{R}^O$ in the robot frame and the object mask $\mathbf{M}_{\mathbf{H}_1}^O$ in the initial human frame, and between the object mask $\mathbf{M}_{\mathbf{H}_1}^O$ in the initial human frame and the object mask $\mathbf{M}_{\mathbf{H}_n}^O$ in the last human frame) and one pair of slot local patches (between the slot $\mathbf{M}_{\mathbf{H}_1}^{S}$ in the initial human frame and the slot $\mathbf{M}_\mathbf{R}^{S_i}$ in the robot view for any slot $S_i$ to place).

\vspace{1mm}
\noindent
{\bf 3D lifting and transformation calculation.}
Using the depth sensing and the camera intrinsic parameters, we can lift all human and robot view images into 3D point cloud observations.
Then, we are able to lift the 2D keypoint correspondences into 3D correspondences.
Equipped with the 3D correspondences, we use Procrustes analysis~\cite{goodall1991procrustes} with RANSAC~\cite{fischler1981random} to calculate three 6-DoF transformation matrices for the aforementioned three local patch pair matchings.
We denote the three computed 6-DoF transformations as $T^O_{\mathbf{R}\rightarrow\mathbf{H}}$, $T^O_{\mathbf{H}}$, and $T^{S_i}_{\mathbf{H}\rightarrow\mathbf{R}}$ respectively.
Fig.~\ref{fig:correlate_robot} (middle right) illustrates their geometric meanings: the object transformation from the robot scene to the human scene at the start of the human video, the transformation applied by the person to the picked object in the human video, and the slot transformation from the human scene to the robot scene.

\vspace{1mm}
\noindent
{\bf Final object placement transformations.}
As clearly illustrated in Fig.~\ref{fig:correlate_robot} (middle right and rightmost), by chaining up the three 6-DoF transformation matrix explained above, we can compute the final desired 6-DoF transformation matrix for the robot to execute in order to transform the pick object $O$ from its initial position to any target slot $i$ in the robot coordinate frame as the following
\vspace{-2mm}
\begin{equation}
\mathbf{T}_i=\mathbf{T}_{\mathbf{H}\rightarrow\mathbf{R}}^{S_i}\mathbf{T}^O_\mathbf{\mathbf{H}}\mathbf{T}^O_{\mathbf{R}\rightarrow\mathbf{H}}.
\end{equation}
\vspace{-4mm}

\vspace{-3mm}
\section{Experiments}
\label{sec:experiment}
\vspace{-2mm}
We propose a new dataset and present an extensive evaluation of our system in Sec.~\ref{sec:system_eval}, where our system, ~\systemName, outperforms the baselines by a large margin. In Sec.~\ref{sec:ablation}, we present an in-depth ablation over~\modelName and additional design choices in~\systemName. In Sec.~\ref{sec:real_robot}, we show that~\systemName is effective with real-world robots. 

 \begin{figure*}[!tbp]
\centering
\includegraphics[width=\linewidth]{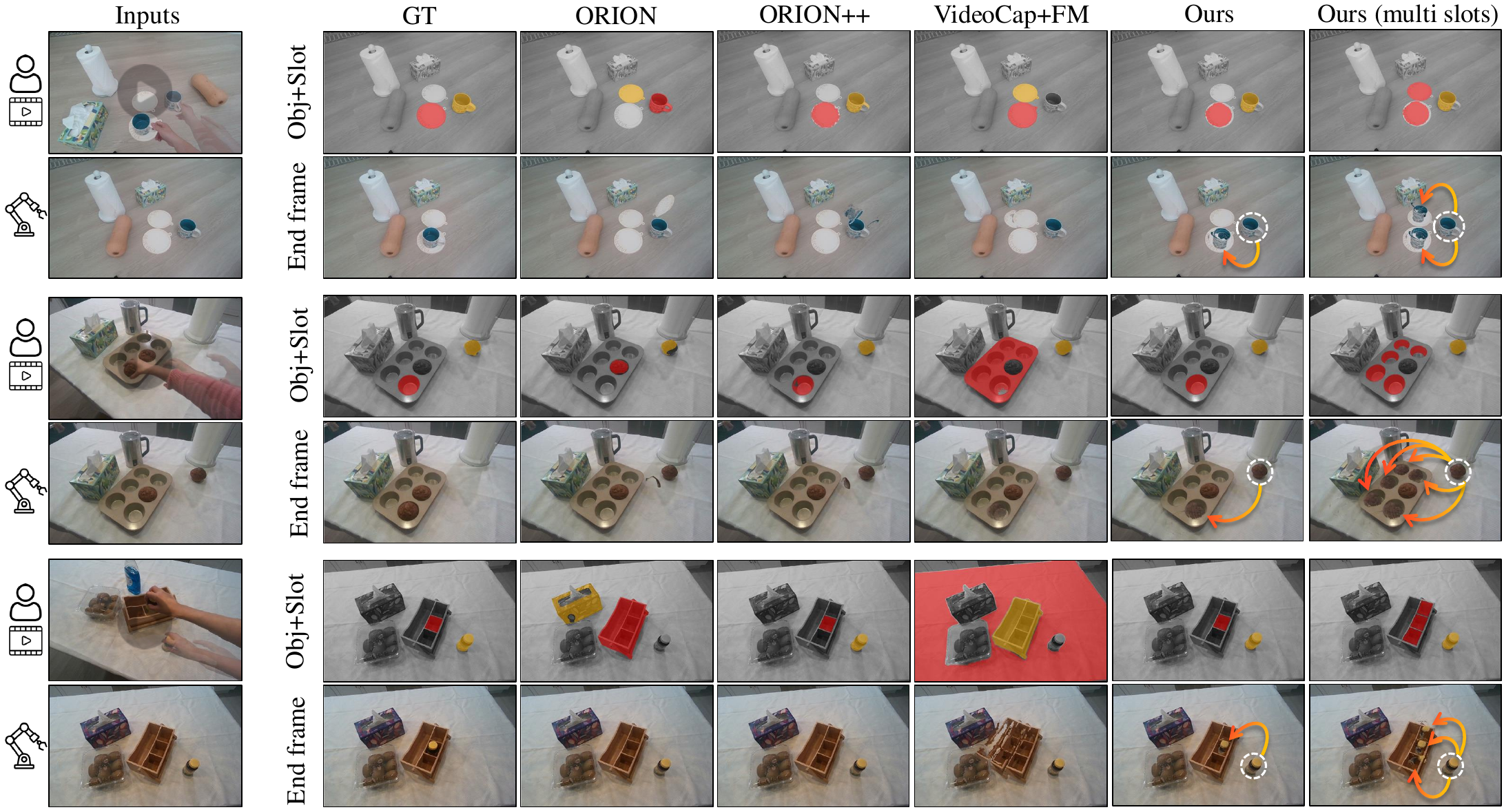}
\vspace{-6mm}
\caption{\textbf{Qualitative Comparison.} We compare our method to baselines and present side-by-side results on three examples. For each example, the first column shows the input human video at the top and robot-view image in the bottom. The top row displays 2D re-identification results (object in \textcolor[rgb]{1,0.69,0}{yellow}, slot in \textcolor[rgb]{1,0,0}{red}), while the bottom row shows 6-DoF relative pose predictions by projecting the object point cloud onto the slots. Unlike the baselines that can only predict one exact slot, our approach can also identify multiple slots. These results clearly demonstrate that our system outperforms the baselines, achieving accurate slot and transformation predictions.
}
\label{fig:vp_results}
\vspace{-3mm}
\end{figure*}

\vspace{-1mm}
\subsection{System Evaluation}
\label{sec:system_eval}
\vspace{-1mm}
Given the novelty of the problem we address, existing evaluation benchmarks are unavailable, and there are no baseline methods with which to make direct comparisons. Consequently, we have curated a dataset comprising real-world videos and established a benchmark specific to this problem by developing suitable baselines and metrics.

\vspace{1mm}
\noindent
{\bf Dataset.}
We collected 288 real-world RGB-D videos spanning 9 different object-in-slot task scenarios. For each scenario, variations were introduced in the background and the inclusion of distractor objects, camera positions, and slot occupancy conditions.
The suite of tasks includes challenging, common daily activities such as \textit{putting bread into a toaster}, \textit{arranging eggs in an egg steamer}, and \textit{setting mugs on coasters}. 
All the objects are unseen to \modelName during training, and 3 out of the 9 tasks encompass previously unseen task categories.
Visualizations of the tasks and their varying settings are provided in the supplementary material.

\vspace{1mm}
\noindent
{\bf Benchmark setup.}
Given that our task necessitates paired data comprising a human demonstration video and a novel image for the robot's view, we construct test pairs by re-pairing the videos in our dataset. Each pair comprises videos depicting the same object being placed into the same slot, albeit with potential variations in background, camera angle, and initial slot occupancy. We designate the first video in each pair as the human demonstration and employ the initial frame of the second video as the robot’s view, as illustrated in Fig.~\ref{fig:vp_results}. During video data collection, we ensure that human hands are absent in the first frame. We generate three distinct test splits, introducing variations in viewpoint (288 video pairs), background (720 video pairs), and slot occupancy (288 video pairs).

\noindent
{\bf Metrics.}
We evaluate the accuracy of the predicted 2D masks and the 6-DoF transformation matrix using five distinct metrics. For the evaluation of 2D masks, we calculate (1) the intersection-over-union (IoU) for the object mask and (2) the IoU of the exact slot mask in the robot view, comparing them to their respective ground-truth masks. 
We assess the accuracy of the 6-DoF transformation by transforming and projecting the object’s point cloud onto the robot view, then measuring (3) the precision of the mask against the ground-truth mask.
For 3D evaluation, we compute (4) the Chamfer Distance and (5) Earth Mover’s Distance~\cite{fan2017point} between the transformed object point cloud and the ground-truth object point cloud at placement. 
If no mask or transformation output is predicted for any method, we use a default empty mask and identity matrix as the fallback predictions.
To establish ground truths, we annotate the 2D masks of the object and exact slot in the start video frame (\ie, robot's view), alongside the object's mask post-placement in the end frame.

\begin{table*}[t]
  \resizebox{\textwidth}{!}{
  \centering
  
  \newcommand{\gto}{~~}
  \begin{tabular}{@{}l|c@{\gto}c@{\gto}c@{\gto}c@{\gto}c|c@{\gto}c@{\gto}c@{\gto}c@{\gto}c|c@{\gto}c@{\gto}c@{\gto}c@{\gto}c@{}}
    \toprule
    \multirow{2}{*}{Method} & \multicolumn{5}{c}{Different view} &  \multicolumn{5}{c}{Different background} & \multicolumn{5}{c}{Different slot occupancy}  \\
    & Obj$\uparrow$ &  Slot$\uparrow$ & Prec.$\uparrow$ & CD$\downarrow$  & EMD$\downarrow$
    & Obj$\uparrow$ &  Slot$\uparrow$ & Prec.$\uparrow$ & CD$\downarrow$ &EMD$\downarrow$ 
    & Obj$\uparrow$ &  Slot$\uparrow$ & Prec.$\uparrow$ & CD$\downarrow$ &EMD$\downarrow$ \\
    \midrule
    ORION~\cite{zhu2024vision}  &0.00  &0.11  &0.41  &0.0949  &0.0552  &0.45  &0.13  &0.00  &0.0932  &0.0540  &0.40 &0.12 &0.82 &0.0952 &0.0559\\
    ORION++~\cite{zhu2024vision}  &10.21 &8.89 &2.10 &0.1058 &0.0584 &7.53 &5.97 &0.75 &0.1113 &0.0597 &8.83 &7.44 &2.95 &0.1021 &0.0576 \\
    CLIPort++~\cite{shridhar2021cliport}  &1.54 &0.47 &18.75 &0.3887 &0.1663 &1.51 &0.35 &1.71 &0.1152 &0.0615 &3.06 &0.34 &12.50 &0.1348 &0.0681 \\
    VideoCap+FMs &2.35 &8.61  &13.45  &0.1918  &0.0987  &2.12  &6.28  &13.61  &0.1743  &0.0917  &2.64  &9.84  &7.25  &0.1508  &0.0837   \\
    Ours                        &\textbf{73.85}  &\textbf{54.37}  &\textbf{36.40}  &\textbf{0.0282}  &\textbf{0.0182}  &\textbf{70.27}  &\textbf{44.70}  &\textbf{25.39}  &\textbf{0.0573}  &\textbf{0.0323}  &\textbf{68.12}  &\textbf{47.04}  &\textbf{30.30}  &\textbf{0.0334}  &\textbf{0.0223}\\
    \bottomrule
  \end{tabular}
  }
    \vspace{-3mm}
\caption{\textbf{Quantitative System Evaluation.} We compare \systemName with baselines and report the 2D detection and the 3D object transformation accuracy: IoU for the object mask prediction ({\it Obj}); IoU for the slot mask prediction ({\it Slot}); mask precision of the predicted object after placement onto the slot projected to the camera plane ({\it Prec.}); and Chamfer distance ({\it CD}) and Earth-Mover distance ({\it EMD}) between the predicted and ground-truth target object point clouds after placement. We evaluate in three different settings with the robot's views having different camera viewpoints, scene backgrounds, and initial states of the placement slot occupancy compared to the input human videos. We find that \systemName substantially outperforms the baselines by large margins across all the metrics in all the three evaluation settings.
}
  \label{tab:vp_comparison}
      \vspace{-4mm}
\end{table*}

\begin{table}[t]
  \resizebox{0.5\textwidth}{!}{
  \centering
  \scriptsize
  \begin{tabular}{@{}l|cc|cc|ccc@{}}
    \toprule
    \multirow{2}{*}{Method} & \multicolumn{2}{c}{Synthetic} &  \multicolumn{2}{c}{Real (seen tasks)} & \multicolumn{2}{c}{Real (unseen tasks)}  \\
     & F1 $\uparrow$ & IoU $\uparrow$
     & F1 $\uparrow$ & IoU $\uparrow$ 
     & F1 $\uparrow$ & IoU $\uparrow$ \\
    \midrule
    Image difference  & 44.10 & 31.34 & 32.69 & 20.04  & 32.20 & 19.53\\
    Change detection~\cite{fang2022changer}  & 2.27 & 1.58  & 0.06 & 0.03 & 6.30 & 4.19 \\
    Object mask       & 60.16 & 48.73 & 65.90 & 52.10 & 58.67 & 42.98\\
    Object-box mask & - & - & 43.39 & 37.74 & 39.37 & 34.65\\
    GPT4o+SAM      & 0.91 & 0.49  & 4.82 & 2.84 & 4.45 & 2.68  \\
    \midrule
    \modelName (end image)      & \textbf{83.44} & 74.57 & 48.83 & 38.68 &38.19 & 28.37\\
    \modelName (ours)       & 82.89 & \textbf{74.62}  & \textbf{73.27} & \textbf{61.59} & \textbf{66.50} & \textbf{54.26} \\
    \bottomrule
  \end{tabular}}
      \vspace{-2mm}
  \caption{\textbf{Slot Segmentation Results.} We compare \modelName against various alternative approaches on slot detection. 
  We evaluate on test synthetic images and our collected real-world images (seen and unseen tasks).
  Dashes note that the method cannot be evaluated for synthetic data given no video inputs.
  We can observe cleary that our \modelName performs the best.
  }
  \label{tab:slotnet}
      \vspace{-6mm}
\end{table}

\vspace{1mm}
\noindent
{\bf Baselines.}
We design four baselines for comparison. 
\begin{itemize}
\item {\it ORION.} Zhu \etal.~\cite{zhu2024vision} perform vision-based human-to-robot imitation learning focused on object-level placement. We adapt ORION to our slot-level setting by providing the required ground-truth object and slot names.
In contrast, our method automatically detects in-contact objects and slots without the need of explicit name inputs. 
\item {\it ORION++.} We leverage the object and slot detection results from \systemName to enhance ORION, thereby establishing a stronger baseline.
\item{\it CLIPort++.} CLIPort~\cite{shridhar2021cliport} is an end-to-end imitation-learning-based language-conditioned policy for tabletop tasks. However, the original method requires videos with action labels for training, whereas ours does not. To construct a comparison, we randomly split the tasks into training and test sets, ensuring that tasks, objects, and scenes are unseen during testing, and use the training split to train CLIPort (more details in Supplementary).
\item {\it VideoCap+FMs.} We test whether our proposed tasks can be effectively solved using state-of-the-art foundation models. We utilize Qwen2VL~\cite{Qwen2VL}, a video captioning model, to summarize the video, and then employ GPT4o~\cite{achiam2023gpt} to identify the object and slot names. Subsequently, we use grounding-SAM2~\cite{ren2024grounding} to generate a bounding box and employ SAM2~\cite{ravi2024sam} to produce the object and slot mask, with the transformation matrix computed using modules in the same way as in our system.
\end{itemize}

\vspace{1mm}
\noindent
{\bf Results.}
Table\ref{tab:vp_comparison} presents a quantitative evaluation comparing \systemName to four baseline methods. The results indicate that \systemName significantly outperforms all baselines by considerable margins. Fig.\ref{fig:vp_results} offers qualitative comparisons, showcasing 2D object and slot mask predictions alongside 3D object transformation estimations. We observe that \systemName generates more accurate 2D object and slot detection results, as baseline methods such as ORION and VideoCap+FMs frequently struggle to describe slot names in natural language for subsequent visual recognition (e.g., incorrectly detecting the entire placement object or table). Additionally, \systemName achieves more precise 3D object transformations compared to baselines like CLIPort, which are primarily designed for top-down 2D predictions.

Lastly, our method can fill multiple slots, whereas other methods generate output for only a single slot. In a subset of the data with ground-truth annotations for multiple slots, \systemName achieves IoU scores of 67.70 and 42.62 for 2D object and slot segmentation. Additionally, it attains scores of 23.14, 0.0575, and 0.0350 for 3D transformation predictions in terms of slot projection precision, Chamfer Distance (CD), and Earth Mover’s Distance (EMD), respectively. These results are comparable to the single-slot placement evaluations reported in Table~\ref{tab:vp_comparison}. See the supplementary materials for further details.

\begin{table}[t]
  \resizebox{0.5\textwidth}{!}{
  \centering
  
  \setlength{\tabcolsep}{2pt}
  \begin{tabular}{@{}l|ccc|ccccc@{}}
    \toprule
   Method (diff. view) & SlotNet & SAM2 & Mast3r
    & Obj $\uparrow$ &  Slot $\uparrow$ & Prec. $\uparrow$ & CD $\downarrow$  & EMD $\downarrow$ 
 \\
    \midrule
    Base design      & \xmark & \xmark & \xmark                           &28.53 &28.23  &24.72  &0.2289  &0.1183\\
    Ours w/o SlotNet  & \xmark & \cmark & \cmark                           &\textbf{73.85}  &29.63  &27.84  &0.0486  &0.0289 \\
    Ours w/o SAM2    & \cmark & \xmark & \cmark         &31.05  &35.98  &23.75  &0.1219  &0.0667 \\
    Ours w/o Mast3r   & \cmark & \cmark & \xmark &\textbf{73.85}  &\textbf{54.37}  &32.74  &0.0321  &0.0205 \\
    Ours             & \cmark & \cmark         & \cmark                   &\textbf{73.85}  &\textbf{54.37}  &\textbf{36.40}  &\textbf{0.0282}  &\textbf{0.0182} \\
    
    \bottomrule
  \end{tabular}
  }
    \vspace{-2mm}
  \caption{
  \textbf{Ablation Study.} All metrics follow Table~\ref{tab:vp_comparison}. 
  Results show all the key modules help.
  See supplementary for the full table.
  }
  \label{tab:vp_ablation}
      \vspace{-4mm}
\end{table}
 \begin{figure*}[!tbp]
\centering
\includegraphics[width=\linewidth]{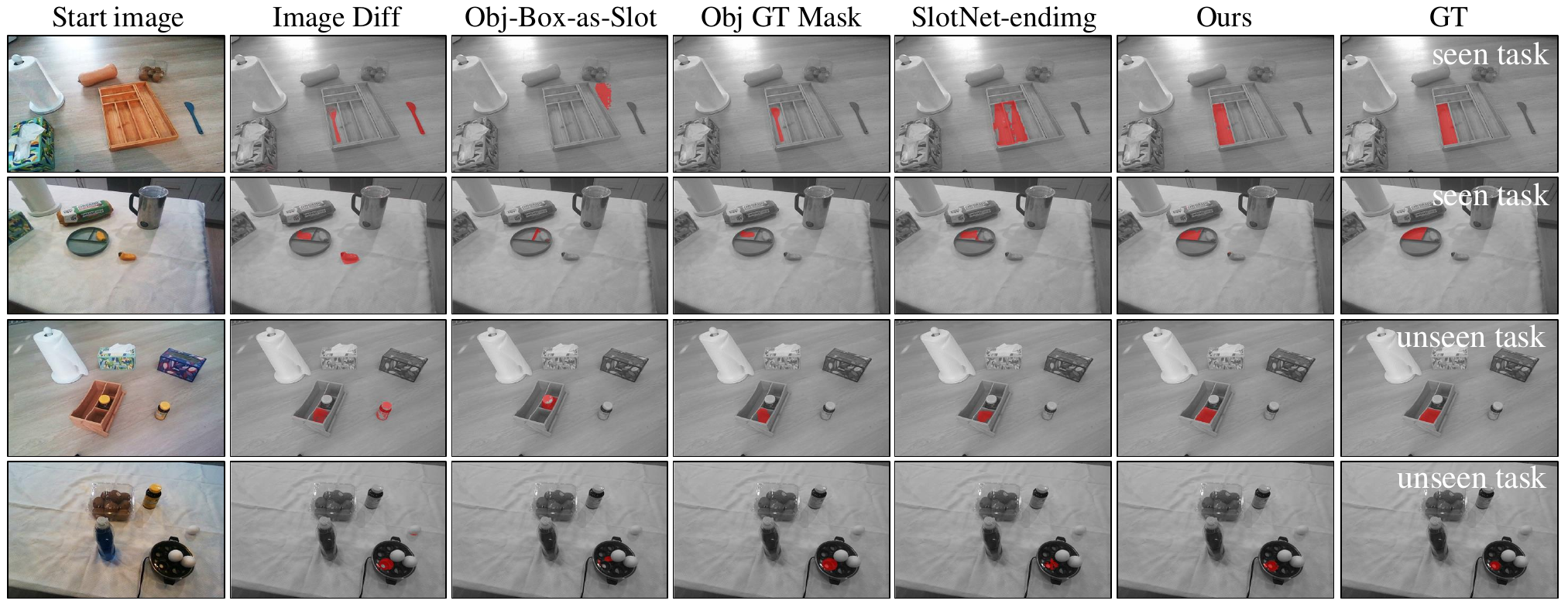}
\vspace{-6mm}
\caption{\textbf{Slot Detection Comparisons.} We compare \modelName to many alternative approaches, and results show that ours performs better.}
\label{fig:slot_results}
\vspace{-4mm}
\end{figure*}

\vspace{-1mm}
\subsection{Ablation Study}
\label{sec:ablation}
We evaluate the effectiveness of \modelName for placement slot detection in Sec.~\ref{sec:slotnet_ablation_study} and additional ablations on other components like object and slot re-identification and key-point matching in Sec.~\ref{sec:more_ablation}. 

\subsubsection{~\modelName Ablations}
\label{sec:slotnet_ablation_study}
\vspace{-1mm}

\vspace{1mm}
\noindent
{\bf Baselines.}
We consider the following alternative and ablation approaches to replace \modelName.

\vspace{-1mm}
\begin{itemize}
\item {\it Image difference.} We use the difference image between the gray-scale start and end frame, then apply thresholding to the difference image to obtain a mask.

\item{\it Change detection.} We use an off-the-shelf change detection model~\cite{fang2022changer} given two frames for the masks.

\item{\it Object mask.} We directly use the ground-truth pick object mask as the slot mask prediction.

\item{\it Object-box mask.} We take the pick object bounding box detected in the tracking procedure and query SAM for a proxy placement slot mask.

\item{\it GPT4o+SAM.} We query GPT4o with start and end frames for slot bounding boxes and query SAM for masks.

\item{\it \modelName (end image).} We use the end frame to replace the difference image as the prompt for \modelName.

\end{itemize}

\noindent
{\bf Benchmark and metrics.}
We use our newly proposed real-world video dataset along with held-out synthetic images for evaluation. For the real images, we evaluate two splits: \emph{seen tasks}, which involve seen object categories during training but novel object instances, and \emph{unseen tasks}, featuring object categories not encountered during training. For each real-world video, we pair the starting and ending frames as input and evaluate predictions against the ground-truth slot mask from the starting frame. The slot mask prediction performance is assessed using IoU and F1 scores.

\vspace{1mm}
\noindent
{\bf Results.}
Table~\ref{tab:slotnet} shows that \modelName, when trained on synthetic data, generalizes effectively to real images, outperforming alternative approaches. Fig.~\ref{fig:slot_results} provides side-by-side comparisons of different methods, revealing that \modelName excels in identifying slot boundaries of various shapes. This underscores the necessity of training a custom model for slot detection and demonstrates that our model is both well-designed and effective.

\vspace{-1mm}
\subsubsection{Other Ablations}
\vspace{-1mm}
\label{sec:more_ablation}
Beyond ablating \modelName, we further validate two additional key design elements in our system: utilizing SAM2 for object and slot re-identification and employing MASt3R for keypoint matching. In the absence of SAM2, we rely on DINOv2 feature similarity between the slot mask and all SAM-generated masks in the robot image. To replace MASt3R, we employ DINOv2 features for Hungarian matching. Table~\ref{tab:vp_ablation} shows the necessity of these modules.

\begin{figure}[!tbp]
\centering
\includegraphics[width=\linewidth]{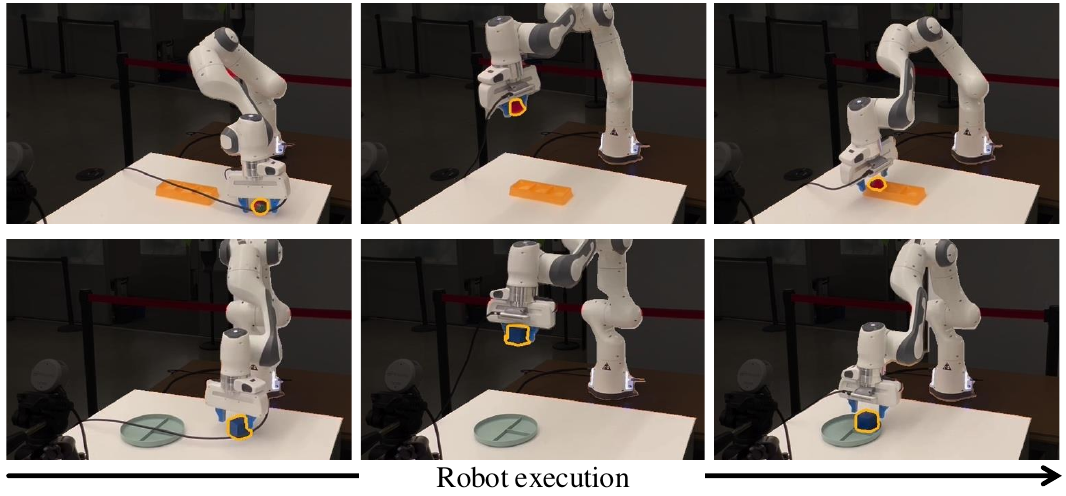}
\vspace{-8mm}
\caption{\textbf{Real-Robot Experiments.} We show real-robot experiments for ``block into a container'' and ``strawberry into an organizer''. See supplementary for videos and more examples.}

\label{fig:robot}
\vspace{-4mm}
\end{figure}

\vspace{-1mm}
\subsection{Real-Robot Experiments}
\label{sec:real_robot}
\vspace{-1mm}
As shown in Fig.~\ref{fig:robot}, we perform real-robot experiments with a Franka robot and show that~\systemName is effective for real robots. The manipulation system employs a wrist-mounted RGB-D camera (Realsense D415) and an external RGB-D camera (Realsense L515). The camera intrinsics and extrinsics relative to the robot are known. The wrist-mounted camera provides observations for \systemName, while the external camera observes the entire scene and aids in planning collision-free trajectories. The system utilizes Contact-Graspnet\cite{sundermeyer2021icra} to generate grasps and plans collision-free trajectories using methods described in \cite{danielczuk2021object, murali2023cabinet}.

\vspace{-2mm}
\section{Conclusion}
\label{sec:conclusion}
\vspace{-2mm}
In this paper, we address the novel problem of slot-level object placement by learning from a single human demonstration video. We propose a modular system to tackle this problem, which operates without requiring additional training video data and features a unique slot-level placement detector. To evaluate the system's performance, we introduce a new benchmark consisting of real-world videos and compare our system against key baseline methods. Our results demonstrate that \systemName outperforms these baselines and functions effectively in real-robot experiments.

\vspace{1mm}
\noindent
{\bf Limitations and future work.} 
Given the novel problem formulation, there is potential for further research in fine-grained slot-level object placement with minimal or no human demonstrations. 
Future work could focus on relaxing current system assumptions, such as the static camera, single-handed interaction, and minimal motion of the placement object. Moreover, advancements in visual foundation models could enhance the robustness of our system, as they play a crucial role in this work.

{
    \small
    \bibliographystyle{ieeenat_fullname}
    \bibliography{main}
}

\end{document}


\maketitle

In this supplementary material, we present real-robot demonstration videos and additional details that could not be included in the main paper due to space constraints.

Firstly, we encourage readers to view the real-robot experiment videos listed below, which demonstrate learning from a single human demonstration:

\begin{itemize}
  \setlength{\itemindent}{0em}
  \item \texttt{block\_in\_organizer.mp4} demos picking the block and placing it into the organizer.
  \item \texttt{egg\_in\_egg\_carton.mp4} demos picking the egg and placing it into the egg carton.
  \item \texttt{muffin\_in\_matt\_tray.mp4} demos picking the muffin and placing it into the matt tray.
  \item \texttt{muffin\_in\_muffin\_tray.mp4} demos picking the muffin and placing it into the muffin tray.
  \item \texttt{strabbery\_in\_organizer.mp4} demos picking the strawberry and placing it into the organizer.
\end{itemize}

Additionally, we provide further details on synthetic data generation (Sec.~\ref{sec:dategen}), visualizations of the evaluation video dataset(Sec.~\ref{sec:eval_data}), implementation details (Sec.~\ref{sec:implementation}) and more qualitative comparisons to the baseline and ablation studies of \systemName and \modelName in Sec.~\ref{sec:slerp_results} and Sec.~\ref{sec:slotnet_results}.

\appendix
\tableofcontents

\section{\modelName Data Generation}
\label{sec:dategen}

\subsection{Details, Statistics and Examples}

\noindent
{\bf Object-with-slots image collection. }  
We collect object-centric images of items with slots by photographing them in everyday environments. These images serve as the \textit{end} object crops, and we generate corresponding \textit{start} images by removing objects, thus creating (\textit{start, end}) image pairs for a pick-and-place action. During data collection, participants were instructed to partially fill the slots. This approach ensures that slots are not entirely empty, facilitating object removal, while avoiding fully filled slots, which can complicate inpainting during object removal.
In total, we have collected 2,138 object-centric images spanning 67 categories, as detailed in Tab.~\ref{tab:datagen_stats}.

\begin{table}
    \resizebox{\linewidth}{!}{
        \begin{tabular}{ c | c | c }
        Task Name  & \#(obj centric) & \#synthetic \\
        \hline
        battery in batter compartment & 20 & 50 \\
        bead in bead organizer & 41 & 4,291 \\
        block in a toy vehicle & 12 & 102 \\
        book in book holder & 22 & 2,207 \\
        bottle in box & 28 & 2927 \\
        bottle in organizer & 87 & 7,810 \\
        bread in toaster & 84 & 3,896 \\
        can in box & 5 & 50 \\
        can on board & 15 & 557 \\
        candle in organizer & 5 & 450 \\
        capsule in tray & 20 & 1,966 \\
        cards in organizer & 5 & 51 \\
        choclate in organier & 22 & 200 \\
        coin in organizer & 6 & 451 \\
        cup in cardboard tray & 101 & 10,541 \\
        cup on board & 5 & 245 \\
        cup on coaster & 5 & 199 \\
        cylinder in organizer & 49 & 1,705 \\
        egg in egg carton & 205 & 36,758 \\
        egg in egg steamer & 13 & 649 \\
        egg in tray & 22 & 2,116 \\
        flower in organier & 15 & 250 \\
        food in organizor & 123 & 9,117 \\
        food in tray & 10 & 1,250 \\
        fruit in box & 5 & 450 \\
        fruit in food organizer & 35 & 3,146 \\
        fruit in organizer & 155 & 11,303 \\
        glass in organizer & 5 & 400 \\
        glass on coaster & 7 & 200 \\
        glove in organizer & 12 & 400 \\
        ice cube in ice tray & 31 & 661 \\
        jewerly in organizer & 18 & 960 \\
        key in organizer & 6 & 900 \\
        lip stick in organizer & 14 & 256 \\
        muffin in muffin tray & 63 & 4,150 \\
        mug on board & 10 & 310 \\
        mug on coaster & 66 & 2,453 \\
        notepad in box & 15 & 901 \\
        peach in box & 12 & 903 \\
        peg in wood base & 8 & 1,413 \\
        pen in basket & 15 & 793 \\
        pen in cup & 20 & 994 \\
        pen in organizer & 103 & 8,690 \\
        pen in pen holder & 76 & 3,305 \\
        pen on book & 30 & 1,335 \\
        pepper in tray & 10 & 492 \\
        pill in pill organizer & 15 & 462 \\
        plant in vase & 12 & 387 \\
        rectangle box in tray & 35 & 366 \\
        tangerine in muffin tray & 24 & 1,700 \\
        tool in organizer & 61 & 400 \\
        toy car in organizer & 14 & 603 \\
        tube in tray & 15 & 1,400 \\
        utensil in bowl & 20 & 850 \\
        utensil in cup & 19 & 2,311 \\
        utensil in utensil organizer & 146 & 14,976 \\
        wood in organizer & 9 & 1,100 \\
        \hline
        \end{tabular}
    }
    \caption{Statistics for object-centric images. This table shows the task names for the object-centric images we collected (col 1), the amount of object-centric images (col 2), and the amount of corresponding generated images (col 3).}
    \label{tab:datagen_stats}
\end{table}

\vspace{1mm}
\noindent
{\bf Pick-object removal.} 
The captured image serves as an object-centric \textit{end} image. To create an object-centric \textit{start} image, where objects are absent from the slot, we employ the SDXL~\cite{von-platen-etal-2022-diffusers} inpainting model to remove the pick-object. For cases where SDXL does not successfully remove the object, we use Cleanup.pictures~\cite{cleanup} in a subsequent round to achieve more effective removal results.

\vspace{1mm}
\noindent
{\bf Slot mask annotation.} 
After obtaining the (\textit{start, end}) object-centric images, we annotate the slot mask by comparing each pair using TORAS~\cite{toras}. We perform mask annotation prior to augmentation and automatically transform these masks onto the canvas, enabling us to generate labels for a large dataset.

\vspace{1mm}
\noindent
{\bf Augmentation with outpainting.} 
For each object-centric slot object, we create a larger scene by first sampling random locations on a 1024x1024 canvas and then applying outpainting with various generated prompts. Using a short category name, such as ``bread in toaster'', we employ Llama~\cite{touvron2023llama} to enrich the text prompts with descriptions of the environments. We apply outpainting to the background following the Hugging Face Outpainting script~\cite{huggingface_outpainting} with enriched texts to create the \textit{end} image. The outpainting script incorporates ControlNet~\cite{zhang2023adding}, SDXL~\cite{von-platen-etal-2022-diffusers}, and ZoeDepth~\cite{bhat2023zoedepth}. Subsequently, the start object crop images and slot masks are subjected to the same transformation to create the outpainted (\textit{start, end}) image pairs and the corresponding ground-truth slot masks. In total, we apply 100 augmentations for each slot on the object-centric image, resulting in 156,000 images for training. Examples of ``egg in egg carton'' augmentation are shown in Fig.~\ref{fig:supp_data_aug}.

\begin{figure}[!tbp]
\centering
\includegraphics[width=0.8\linewidth]{figures/supp/data_aug.pdf}
\caption{\textbf{Data augmentation.} We show 9 synthetic data augmentation examples for the ``egg in egg carton'' task. In training data generation, we use x100 augmentation for each slot in the object-centric images.}
\label{fig:supp_data_aug}
\vspace{-3mm}
\end{figure}

\subsection{Synthetic Data Generation Process}
We present examples of our synthetic generation process in Tab.~\ref{tab:datagen_example1} and Tab.~\ref{tab:datagen_example2}. Each row in the table, from left to right, illustrates the step-by-step process for one example.

Beginning with a task name (col 1) and an object-centric image (col 3), we first enhance the description using Llama~\cite{touvron2023llama} (col 2). Next, we perform object removal from the slots through two rounds of image inpainting using SDXL~\cite{von-platen-etal-2022-diffusers} and Clean.pictures~\cite{cleanup} (col 4) to ensure more effective results. We annotate the slot mask on the object-centric image utilizing the Toronto Annotation Suite (TORAS)~\cite{toras} (col 5).

For generating diverse daily backgrounds, we transform the object-centric image (col 3) onto a 1024x1024 canvas and outpaint the background using the Hugging Face Outpainting script~\cite{huggingface_outpainting} with the enriched text generated by Llama~\cite{touvron2023llama} (col 2) as a prompt to create the \textit{end} image. Finally, the \textit{start} image and ground-truth mask undergo the same transformation onto a 1024x1024 canvas to produce the (\textit{start}, \textit{end}, mask) triplet samples for training.

\begin{table*}[t]
    \vspace{-5mm}
    \resizebox{\textwidth}{0.47\textheight}{
    \begin{tabular}{|p{1cm}|p{2cm}|c|c|c|c|c|c|}
        \hline
        Task & Enrich text & Obj-centic img & Obj removal & Annotation & \textit {Start} &  \textit {End} & GT mask \\
        \hline
        \small egg in egg carton 
        & \tiny {'top': 'egg', 'bottom': 'egg carton', 'style': 'High-quality and photorealistic, modern, bright, morning, 45-degree look-down view, 2020s', 'objects': 'A large, brown egg with a smooth, glossy texture, positioned in the center of a compact, rectangular egg carton with a textured, ribbed surface, nestled snugly in its compartment.', 'background': 'A realistic, everyday kitchen scene, with a plain, white countertop, a glass jar of jam, a toaster, a stack of plates, a warm, golden-brown wood countertop, and soft, creamy white walls.'}
        & {\includegraphics[scale=0.12, valign=t]{figures/supp/datagen_examples/egg_in_egg_carton__0002.jpg}}
        & \includegraphics[scale=0.12, valign=t]{figures/supp/datagen_examples/egg_in_egg_carton__0002__pickoj_00.jpg__1__0.38.jpg} 
        & \includegraphics[scale=0.12, valign=t]{figures/supp/datagen_examples/egg_in_egg_carton__0002__pickoj_00_mask.png}  
        & \includegraphics[scale=0.12, valign=t]{figures/supp/datagen_examples/egg_in_egg_carton__0002__pickoj_00.jpg__attempt_58__0000__start.jpg}
        & \includegraphics[scale=0.12, valign=t]{figures/supp/datagen_examples/egg_in_egg_carton__0002__pickoj_00.jpg__attempt_58__0000__end.jpg} 
        & \includegraphics[scale=0.12, valign=t]{figures/supp/datagen_examples/egg_in_egg_carton__0002__pickoj_00.jpg__attempt_58__0000__mask.png} \\
        
        \hline
        \small bottle in box 
        & \tiny {{'top': 'box', 'bottom': 'bottle is inside the box', 'style': '2020s contemporary style with bright and luminous brightness, set in the morning time with an eye-level shot', 'objects': 'The bottle is a sleek, modern glass bottle with a smooth texture and a minimalist design. It has a simple label with a subtle texture. The box is a sturdy, corrugated cardboard box with a textured surface and a classic design.', 'background': 'A realistic, everyday living room scene with a few daily supplies around the box and bottle, including a coffee table with a vase, a few books, and a newspaper in the background, with neutral-colored walls and a light-colored rug on the floor.'}}
        & {\includegraphics[scale=0.12, valign=t]{figures/supp/datagen_examples/bottle_in_box__0004.jpg}}
        & \includegraphics[scale=0.12, valign=t]{figures/supp/datagen_examples/bottle_in_box__0004__pickoj_03.jpg__1__0.11.jpg} 
        & \includegraphics[scale=0.12, valign=t]{figures/supp/datagen_examples/bottle_in_box__0004__pickoj_03_mask.png}
        & \includegraphics[scale=0.12, valign=t]{figures/supp/datagen_examples/bottle_in_box__0004__pickoj_03.jpg__attempt_95__0000__start.jpg}
        & \includegraphics[scale=0.12, valign=t]{figures/supp/datagen_examples/bottle_in_box__0004__pickoj_03.jpg__attempt_95__0000__end.jpg} 
        & \includegraphics[scale=0.12, valign=t]{figures/supp/datagen_examples/bottle_in_box__0004__pickoj_03.jpg__attempt_95__0000__mask.png} \\

        \hline
        \small bread in toaster 
        & \tiny {{'top': 'toaster', 'bottom': 'bread', 'style': 'A 2020s, modern, bright, and luminous style with a 45-degree look-down view, capturing the morning scene with a sense of coziness and warmth.', 'objects': "The toaster, a sleek and modern appliance with a stainless steel finish and a compact design, sits proudly on the countertop. The bread, a soft and fluffy white loaf, is nestled inside the toaster's slots, awaiting its turn to be toasted.", 'background': 'The countertop is a warm and inviting beige color, with a few subtle scratches and scuffs to give it a sense of character. A few daily supplies are scattered around the toaster, including a jar of jam, a carton of eggs, and a loaf of bread, adding to the sense of a busy morning routine.'}}
        & {\includegraphics[scale=0.12, valign=t]{figures/supp/datagen_examples/bread_in_toaster__0022.jpg}}
        & \includegraphics[scale=0.12, valign=t]{figures/supp/datagen_examples/bread_in_toaster__0022__pickoj_01.jpg__1__0.56.jpg} 
        & \includegraphics[scale=0.12, valign=t]{figures/supp/datagen_examples/bread_in_toaster__0022__pickoj_01_mask.png}  
        & \includegraphics[scale=0.12, valign=t]{figures/supp/datagen_examples/bread_in_toaster__0022__pickoj_01.jpg__attempt_22__0022__start.jpg}
        & \includegraphics[scale=0.12, valign=t]{figures/supp/datagen_examples/bread_in_toaster__0022__pickoj_01.jpg__attempt_22__0022__end.jpg} 
        & \includegraphics[scale=0.12, valign=t]{figures/supp/datagen_examples/bread_in_toaster__0022__pickoj_01.jpg__attempt_22__0022__mask.png} \\

        \hline
        \small egg in egg steamer
        & \tiny {{'top': 'egg', 'bottom': 'egg steamer', 'style': 'high-quality and photorealistic, modern, 2020s, bright, noon afternoon, 45-degree look-down view', 'objects': 'The egg is a large, brown egg with a smooth, glossy texture, placed in the egg steamer basket. The egg steamer is a stainless steel, modern kitchen appliance with a sleek and shiny surface, and a simple, geometric design.', 'background': 'A modern kitchen with a white countertop, a stainless steel sink, and a few kitchen utensils and appliances around. A few kitchen towels and a fruit bowl are nearby, adding to the daily living environment.'}}
        & {\includegraphics[scale=0.12, valign=t]{figures/supp/datagen_examples/egg_in_egg_steamer__0001.jpg}}
        & \includegraphics[scale=0.12, valign=t]{figures/supp/datagen_examples/egg_in_egg_steamer__0001__pickoj_00.jpg__1__0.40.jpg} 
        & \includegraphics[scale=0.12, valign=t]{figures/supp/datagen_examples/egg_in_egg_steamer__0001__pickoj_00_mask.png}  
        & \includegraphics[scale=0.12, valign=t]{figures/supp/datagen_examples/egg_in_egg_steamer__0001__pickoj_00.jpg__attempt_80__0006__start.jpg}
        & \includegraphics[scale=0.12, valign=t]{figures/supp/datagen_examples/egg_in_egg_steamer__0001__pickoj_00.jpg__attempt_80__0006__end.jpg} 
        & \includegraphics[scale=0.12, valign=t]{figures/supp/datagen_examples/egg_in_egg_steamer__0001__pickoj_00.jpg__attempt_80__0006__mask.png} \\

        \hline
        \small fruit in food organizer
        & \tiny {{{'top': 'fruit', 'bottom': 'food organizer', 'style': 'A contemporary 2020s style with a bright and luminous ambiance, a 45-degree look-down view, and a medium shot of the objects to capture both the fruit and the organizer in detail.', 'objects': "The fruit, a vibrant and juicy apple, is placed neatly in the organizer's container, showcasing its natural texture and color. The organizer itself has a sleek and modern design, with a matte finish and a subtle sheen to it. The material appears to be a high-quality plastic, with a slight give when touched.", 'background': "The background is a realistic representation of a modern kitchen, with a plain white wall behind the fruit and the organizer. There's a wooden kitchen table to the left, with a few utensils and a water bottle scattered around it. A window is visible in the background, with a glimpse of a cityscape outside."}}}
        & {\includegraphics[scale=0.12, valign=t]{figures/supp/datagen_examples/fruit_in_food_organizer__0034.jpg}}
        & \includegraphics[scale=0.12, valign=t]{figures/supp/datagen_examples/fruit_in_food_organizer__0034__pickoj_00.jpg__1__0.56.jpg} 
        & \includegraphics[scale=0.12, valign=t]{figures/supp/datagen_examples/fruit_in_food_organizer__0034__pickoj_00_mask.png}  
        & \includegraphics[scale=0.12, valign=t]{figures/supp/datagen_examples/fruit_in_food_organizer__0034__pickoj_00.jpg__attempt_19__0020__start.jpg}
        & \includegraphics[scale=0.12, valign=t]{figures/supp/datagen_examples/fruit_in_food_organizer__0034__pickoj_00.jpg__attempt_19__0020__end.jpg} 
        & \includegraphics[scale=0.12, valign=t]{figures/supp/datagen_examples/fruit_in_food_organizer__0034__pickoj_00.jpg__attempt_19__0020__mask.png} \\

        \hline
        \small muffin in muffin tray
        & \tiny {{'top': 'muffin', 'bottom': 'muffin tray', 'style': 'A high-quality, photorealistic style with a 2020s modern aesthetic, featuring a bright and cozy atmosphere, a 45-degree look-down view, and a medium shot distance.', 'objects': 'The muffin in the muffin tray is a delicious, golden-brown treat with a soft, fluffy texture and a crispy edge. The muffin tray, made of durable, stainless steel, has a sleek, modern design with a slight sheen to it, giving it a premium look.', 'background': 'The background is a realistic, everyday kitchen scene, complete with a stainless steel refrigerator, a modern kitchen island, and a few kitchen utensils scattered about.'}}
        & {\includegraphics[scale=0.12, valign=t]{figures/supp/datagen_examples/muffin_in_muffin_tray__0006.jpg}}
        & \includegraphics[scale=0.12, valign=t]{figures/supp/datagen_examples/muffin_in_muffin_tray__0006__pickoj_02.jpg__1__0.26.jpg} 
        & \includegraphics[scale=0.12, valign=t]{figures/supp/datagen_examples/muffin_in_muffin_tray__0006__pickoj_02_mask.png}  
        & \includegraphics[scale=0.12, valign=t]{figures/supp/datagen_examples/muffin_in_muffin_tray__0006__pickoj_02.jpg__attempt_99__0072__start.jpg}
        & \includegraphics[scale=0.12, valign=t]{figures/supp/datagen_examples/muffin_in_muffin_tray__0006__pickoj_02.jpg__attempt_99__0072__end.jpg} 
        & \includegraphics[scale=0.12, valign=t]{figures/supp/datagen_examples/muffin_in_muffin_tray__0006__pickoj_02.jpg__attempt_99__0072__mask.png} \\

        \hline
    \end{tabular}
    }
    \caption{Data generation process (Part 1/2). Start from a task name (col 1) and an object-centric image (col 3), we first get an enriched detailed description using~\cite{touvron2023llama} (col 2), then remove one object from the slots (col 4) and annotate the slot mask on the object-centric image (col 5). To put the object-centric image into various daily backgrounds, we transform the object-centric image (col 3) onto a 1024x1024 canvas and outpaint the background with the enriched text (col 2) to create \textit{End}. Finally, \textit{Start} and GT mask follow the same transformation onto the canvas.}
    \label{tab:datagen_example1}
\end{table*}

\begin{table*}
    \vspace{-5mm}
    \resizebox{\textwidth}{!}{
    \begin{tabular}{|p{1cm}|p{2cm}|c|c|c|c|c|c|}
        \hline
        Task & Enrich text & Obj-centic img & Obj removal & Annotation & \textit {Start} &  \textit {End} & GT mask \\
        \hline
        \small mug on coaster
        & \tiny {{'top': 'mug', 'bottom': 'coaster', 'style': 'High-quality and photorealistic, 2020s, modern, bright, morning time, 45-degree look-down view', 'objects': 'A ceramic mug with a textured, matte finish, having a simple yet elegant design with a subtle sheen to it, and a wooden coaster with a natural, rustic texture, having a subtle wood grain pattern', 'background': 'A realistic, real-world scene of a kitchen table with a few daily supplies around, such as a jar of coffee beans, a sugar bowl, and a few coffee cups, with a wooden table finish and light, neutral-colored walls'}}
        & {\includegraphics[scale=0.12, valign=t]{figures/supp/datagen_examples/mug_on_coaster__0010.jpg}}
        & \includegraphics[scale=0.12, valign=t]{figures/supp/datagen_examples/mug_on_coaster__0010__pickoj_00.jpg__1__0.44.jpg} 
        & \includegraphics[scale=0.12, valign=t]{figures/supp/datagen_examples/mug_on_coaster__0010__pickoj_00_mask.png}  
        & \includegraphics[scale=0.12, valign=t]{figures/supp/datagen_examples/mug_on_coaster__0010__pickoj_00.jpg__attempt_79__0000__start.jpg}
        & \includegraphics[scale=0.12, valign=t]{figures/supp/datagen_examples/mug_on_coaster__0010__pickoj_00.jpg__attempt_79__0000__end.jpg} 
        & \includegraphics[scale=0.12, valign=t]{figures/supp/datagen_examples/mug_on_coaster__0010__pickoj_00.jpg__attempt_79__0000__mask.png} \\

        \hline
        \small peach in box
        & \tiny {{'top': 'peach', 'bottom': 'box', 'style': 'mid-century modern, high-quality and photorealistic, bright and cozy, morning, 45-degree look-down view', 'objects': 'The peach has a smooth, juicy texture, and a vibrant orange color. The box has a wooden texture, with a natural wood grain pattern, and a soft brown color.', 'background': 'A realistic kitchen scene, with a wooden table and chairs, and a few daily supplies such as a toaster, a coffee maker, and a vase with fresh flowers.'}}
        & {\includegraphics[scale=0.12, valign=t]{figures/supp/datagen_examples/peach_in_box__0002.jpg}}
        & \includegraphics[scale=0.12, valign=t]{figures/supp/datagen_examples/peach_in_box__0002__pickoj_03.jpg__1__0.57.jpg} 
        & \includegraphics[scale=0.12, valign=t]{figures/supp/datagen_examples/peach_in_box__0002__pickoj_03_mask.png}  
        & \includegraphics[scale=0.12, valign=t]{figures/supp/datagen_examples/peach_in_box__0002__pickoj_03.jpg__attempt_1__0002__start.jpg}
        & \includegraphics[scale=0.12, valign=t]{figures/supp/datagen_examples/peach_in_box__0002__pickoj_03.jpg__attempt_1__0002__end.jpg} 
        & \includegraphics[scale=0.12, valign=t]{figures/supp/datagen_examples/peach_in_box__0002__pickoj_03.jpg__attempt_1__0002__mask.png} \\

        \hline
        \small pen in organizer
        & \tiny {{'top': 'pen', 'bottom': 'organizer', 'style': 'high-quality and photorealistic, modern European-style, bright and cozy, morning, 45-degree look-down view', 'objects': 'The pen is a sleek and modern writing instrument with a silver finish and a rubberized grip. The organizer is a compact and functional desk accessory with a wooden base and a mesh pocket.', 'background': 'The background is a realistic depiction of a modern home office, with a wooden desk and a comfortable office chair. There are several daily supplies around the organizer, including a notebook, a stapler, and a cup of coffee.'}}
        & {\includegraphics[scale=0.12, valign=t]{figures/supp/datagen_examples/pen_in_organizer__0002.jpg}}
        & \includegraphics[scale=0.12, valign=t]{figures/supp/datagen_examples/pen_in_organizer__0002__pickoj_01.jpg__1__0.22.jpg} 
        & \includegraphics[scale=0.12, valign=t]{figures/supp/datagen_examples/pen_in_organizer__0002__pickoj_01_mask.png}  
        & \includegraphics[scale=0.12, valign=t]{figures/supp/datagen_examples/pen_in_organizer__0002__pickoj_01.jpg__attempt_76__0050__start.jpg}
        & \includegraphics[scale=0.12, valign=t]{figures/supp/datagen_examples/pen_in_organizer__0002__pickoj_01.jpg__attempt_76__0050__end.jpg} 
        & \includegraphics[scale=0.12, valign=t]{figures/supp/datagen_examples/pen_in_organizer__0002__pickoj_01.jpg__attempt_76__0050__mask.png} \\

        \hline
        \small pepper in tray
        & \tiny {{'top': 'pepper', 'bottom': 'tray', 'style': 'high-quality and photorealistic, 2020s, contemporary, bright, luminous, morning, 45-degree look-down view', 'objects': 'a medium-sized, green bell pepper with a glossy texture, placed in the center of a rectangular, stainless steel tray with a smooth, matte finish', 'background': 'a wooden kitchen table with a few daily supplies, a warm, beige-colored kitchen wall, cabinets, and a window that lets in natural light'}}
        & {\includegraphics[scale=0.12, valign=t]{figures/supp/datagen_examples/pepper_in_tray__0002.jpg}}
        & \includegraphics[scale=0.12, valign=t]{figures/supp/datagen_examples/pepper_in_tray__0002__pickoj_00.jpg__1__0.22.jpg} 
        & \includegraphics[scale=0.12, valign=t]{figures/supp/datagen_examples/pepper_in_tray__0002__pickoj_00_mask.png}  
        & \includegraphics[scale=0.12, valign=t]{figures/supp/datagen_examples/pepper_in_tray__0002__pickoj_00.jpg__attempt_81__0000__start.jpg}
        & \includegraphics[scale=0.12, valign=t]{figures/supp/datagen_examples/pepper_in_tray__0002__pickoj_00.jpg__attempt_81__0000__end.jpg} 
        & \includegraphics[scale=0.12, valign=t]{figures/supp/datagen_examples/pepper_in_tray__0002__pickoj_00.jpg__attempt_81__0000__mask.png} \\

        \hline
        \small pill in pill organizer
        & \tiny {{'top': 'pill', 'bottom': 'pill organizer', 'style': 'High-quality, photorealistic style, 2020s, modern, bright, morning, 45-degree look-down view', 'objects': "A small, round, and white pill with a smooth texture is centered in the image, and it's a medium shot or medium range. The pill is placed inside a small, rectangular, and plastic pill organizer with a grid pattern on the inside. The pill organizer has a sleek and modern design, with a bright and shiny surface.", 'background': 'A realistic, real-world scene, set in a modern kitchen or bathroom. The scene is bright and well-lit, with a 45-degree look-down view from above. The pill organizer is placed on a countertop, surrounded by other daily supplies such as a toothbrush, toothpaste, and a small trash can. The countertop is made of a modern material, such as granite or quartz, and has a few small appliances and gadgets nearby. The walls are painted a light color, and there is a window in the background, allowing natural light to pour in.'}}
        & {\includegraphics[scale=0.12, valign=t]{figures/supp/datagen_examples/pill_in_pill_organizer__0013.jpg}}
        & \includegraphics[scale=0.12, valign=t]{figures/supp/datagen_examples/pill_in_pill_organizer__0013__pickoj_03.jpg__1__0.00.jpg} 
        & \includegraphics[scale=0.12, valign=t]{figures/supp/datagen_examples/pill_in_pill_organizer__0013__pickoj_03_mask.png}  
        & \includegraphics[scale=0.12, valign=t]{figures/supp/datagen_examples/pill_in_pill_organizer__0013__pickoj_03.jpg__attempt_28__0000__start.jpg}
        & \includegraphics[scale=0.12, valign=t]{figures/supp/datagen_examples/pill_in_pill_organizer__0013__pickoj_03.jpg__attempt_28__0000__end.jpg} 
        & \includegraphics[scale=0.12, valign=t]{figures/supp/datagen_examples/pill_in_pill_organizer__0013__pickoj_03.jpg__attempt_28__0000__mask.png} \\

        \hline
        \small utensil in utensil organizer
        & \tiny {{'top': 'utensil', 'bottom': 'utensil organizer', 'style': 'high-quality and photorealistic, modern style from the 2020s, bright brightness, morning time, 45-degree look-down view, ambient light', 'objects': 'The utensil has a smooth and rounded handle, made of stainless steel with a sleek and shiny finish. The utensil organizer has a grid-like structure with small compartments for each utensil, also made of stainless steel.', 'background': 'A realistic representation of a modern kitchen, with a wooden countertop, a stainless steel refrigerator, and a few daily supplies such as a coffee maker and a toaster.'}}
        & {\includegraphics[scale=0.12, valign=t]{figures/supp/datagen_examples/utensil_in_utensil_organizer__0006.jpg}}
        & \includegraphics[scale=0.12, valign=t]{figures/supp/datagen_examples/utensil_in_utensil_organizer__0006__pickoj_01.jpg__1__0.24.jpg} 
        & \includegraphics[scale=0.12, valign=t]{figures/supp/datagen_examples/utensil_in_utensil_organizer__0006__pickoj_01_mask.png}  
        & \includegraphics[scale=0.12, valign=t]{figures/supp/datagen_examples/utensil_in_utensil_organizer__0006__pickoj_01.jpg__attempt_99__0000__start.jpg}
        & \includegraphics[scale=0.12, valign=t]{figures/supp/datagen_examples/utensil_in_utensil_organizer__0006__pickoj_01.jpg__attempt_99__0000__end.jpg} 
        & \includegraphics[scale=0.12, valign=t]{figures/supp/datagen_examples/utensil_in_utensil_organizer__0006__pickoj_01.jpg__attempt_99__0000__mask.png} \\

    \hline
        
    \end{tabular}
    }
    \vspace{-2mm}
    \caption{Data generation process (Part 2/2 cont.). Start from a task name (col 1) and an object-centric image (col 3), we first get an enriched detailed description using~\cite{touvron2023llama} (col 2), then remove one object from the slots (col 4) and annotate the slot mask on the object-centric image (col 5). To put the object-centric image into various daily backgrounds, we transform the object-centric image (col 3) onto a 1024x1024 canvas and outpaint the background with the enriched text (col 2) to create \textit{End}. Finally, \textit{Start} and GT mask follow the same transformation onto the canvas.}
    \label{tab:datagen_example2}
\end{table*}


\subsection{Limitations}



Our data generation pipeline for creating (\textit{start}, \textit{end}) pairs with minimal data collection and annotation is highly efficient, enabling us to train SAM with manageable effort.

Despite its efficiency, the generated images have certain limitations. First, the diversity of the generated image styles is insufficient. Although we employ Llama to enrich text prompts for backgrounds and request diverse styles, the generated images lack sufficient diversity and realism. Second, the generated images sometimes defy physical plausibility. In the process of outpainting the object-centric images, we sample random locations and rotations on the canvas. Occasionally, these locations or rotations are challenging to outpaint while adhering to physical principles, resulting in images that appear unrealistic.

The purpose of outpainting data augmentation is to generate more varied backgrounds, thereby aiding \modelName in achieving generalization. In this context, the appearance of the background does not adversely affect our use case. We anticipate that advances in generative AI models will mitigate these limitations.
\section{Evaluation Videos}
\label{sec:eval_data}

In this paper, we developed a pick-and-place video dataset containing 288 videos spanning 9 tasks for evaluating both~\modelName and~\systemName.

For~\modelName evaluation, we utilize the \textit{start} and \textit{end} images from each video. Among the 9 tasks, 3 are categorized as unseen for~\modelName evaluation: ``bottle in organizer``, ``cup on saucer'', and ``egg in egg steamer''.

For~\systemName evaluation, we select two videos of the same task, one serving as the human demonstration and the other as the robot-view video. The robot-view video has its final frame depicting what the robot's view will look like upon completion of the action, providing the ground-truth \textit{end} frame appearance. The three different settings we use to pair the videos are:
(1) different views (captured from different camera angles), (2) different backgrounds (captured in different environments with setups on distinct tables), and (3) different slot occlusions (captured with varying slot occlusions, where one video has all other slots empty while the other has some slots filled).

In this section, we provide visualizations of the 9 tasks and 3 settings in Tab.~\ref{tab:eval_video_task} and Tab.~\ref{tab:eval_video_setting}.

\subsection{Visualization of 9 tasks}
We show one video for each of the 9 tasks in Tab.~\ref{tab:eval_video_task}. For each video, we out the (\textit{start}, \textit{pick}, \textit{place}, \textit{end}) frames to illustrate the action.

\subsection{Visualization of 3 settings}
We show two example videos for each of the 3 settings (different views, different backgrounds, and different slot occlusions) in Tab.~\ref{tab:eval_video_setting}.

\begin{table*}
    \resizebox{\textwidth}{!}{
    \begin{tabular}{|p{0.5cm}|p{1.7cm}|p{0.8cm}|c|c|c|c|c|c|}
        
        \hline
        No.  & Task & \#Total & Start Frame & Pick Frame & Place Frame &  End Frame \\
        
        \hline
        1 & bread in toaster & 32
        & \includegraphics[scale=0.12, valign=t]{figures/supp/start_pick_place_end/o0-b0-s3-v0/color_000000.jpg}
        & \includegraphics[scale=0.12, valign=t]{figures/supp/start_pick_place_end/o0-b0-s3-v0/color_000040.jpg}
        & \includegraphics[scale=0.12, valign=t]{figures/supp/start_pick_place_end/o0-b0-s3-v0/color_000113.jpg}
        & \includegraphics[scale=0.12, valign=t]{figures/supp/start_pick_place_end/o0-b0-s3-v0/color_000171.jpg}
        \\

        \hline
        2 & bottle in organizer & 32
        & \includegraphics[scale=0.12, valign=t]{figures/supp/start_pick_place_end/o2-b1-s1-v1/color_000000.jpg}
        & \includegraphics[scale=0.12, valign=t]{figures/supp/start_pick_place_end/o2-b1-s1-v1/color_000026.jpg}
        & \includegraphics[scale=0.12, valign=t]{figures/supp/start_pick_place_end/o2-b1-s1-v1/color_000052.jpg}
        & \includegraphics[scale=0.12, valign=t]{figures/supp/start_pick_place_end/o2-b1-s1-v1/color_000102.jpg}\\

        \hline
        3 & fruit in tray & 32
        & \includegraphics[scale=0.12, valign=t]{figures/supp/start_pick_place_end/o4-b0-s1-v0/color_000000.jpg}
        & \includegraphics[scale=0.12, valign=t]{figures/supp/start_pick_place_end/o4-b0-s1-v0/color_000023.jpg}
        & \includegraphics[scale=0.12, valign=t]{figures/supp/start_pick_place_end/o4-b0-s1-v0/color_000061.jpg}
        & \includegraphics[scale=0.12, valign=t]{figures/supp/start_pick_place_end/o4-b0-s1-v0/color_000128.jpg}\\

        \hline
        4 & cup in saucer & 32
        & \includegraphics[scale=0.12, valign=t]{figures/supp/start_pick_place_end/o6-b3-s1-v1/color_000000.jpg}
        & \includegraphics[scale=0.12, valign=t]{figures/supp/start_pick_place_end/o6-b3-s1-v1/color_000036.jpg}
        & \includegraphics[scale=0.12, valign=t]{figures/supp/start_pick_place_end/o6-b3-s1-v1/color_000114.jpg}
        & \includegraphics[scale=0.12, valign=t]{figures/supp/start_pick_place_end/o6-b3-s1-v1/color_000160.jpg}\\

        \hline
        5 & mug in coaster & 32
        & \includegraphics[scale=0.12, valign=t]{figures/supp/start_pick_place_end/o8-b1-s1-v1/color_000000.jpg}
        & \includegraphics[scale=0.12, valign=t]{figures/supp/start_pick_place_end/o8-b1-s1-v1/color_000023.jpg}
        & \includegraphics[scale=0.12, valign=t]{figures/supp/start_pick_place_end/o8-b1-s1-v1/color_000066.jpg}
        & \includegraphics[scale=0.12, valign=t]{figures/supp/start_pick_place_end/o8-b1-s1-v1/color_000089.jpg}\\

        \hline
        6 & egg in egg steamer & 32
        & \includegraphics[scale=0.12, valign=t]{figures/supp/start_pick_place_end/o9-b1-s3-v0/color_000000.jpg}
        & \includegraphics[scale=0.12, valign=t]{figures/supp/start_pick_place_end/o9-b1-s3-v0/color_000025.jpg}
        & \includegraphics[scale=0.12, valign=t]{figures/supp/start_pick_place_end/o9-b1-s3-v0/color_000058.jpg}
        & \includegraphics[scale=0.12, valign=t]{figures/supp/start_pick_place_end/o9-b1-s3-v0/color_000090.jpg}\\

        \hline
        7 & fruit in organizer & 32
        & \includegraphics[scale=0.12, valign=t]{figures/supp/start_pick_place_end/o10-b1-s1-v1/color_000000.jpg}
        & \includegraphics[scale=0.12, valign=t]{figures/supp/start_pick_place_end/o10-b1-s1-v1/color_000024.jpg}
        & \includegraphics[scale=0.12, valign=t]{figures/supp/start_pick_place_end/o10-b1-s1-v1/color_000057.jpg}
        & \includegraphics[scale=0.12, valign=t]{figures/supp/start_pick_place_end/o10-b1-s1-v1/color_000078.jpg}\\

        \hline
        8 & utensil in utensil organizer & 32
        & \includegraphics[scale=0.12, valign=t]{figures/supp/start_pick_place_end/o11-b1-s2-v0/color_000000.jpg}
        & \includegraphics[scale=0.12, valign=t]{figures/supp/start_pick_place_end/o11-b1-s2-v0/color_000022.jpg}
        & \includegraphics[scale=0.12, valign=t]{figures/supp/start_pick_place_end/o11-b1-s2-v0/color_000068.jpg}
        & \includegraphics[scale=0.12, valign=t]{figures/supp/start_pick_place_end/o11-b1-s2-v0/color_000117.jpg}\\

        \hline
        9 & muffin in muffin tray & 32
        & \includegraphics[scale=0.12, valign=t]{figures/supp/start_pick_place_end/o12-b2-s2-v0/color_000000.jpg}
        & \includegraphics[scale=0.12, valign=t]{figures/supp/start_pick_place_end/o12-b2-s2-v0/color_000031.jpg}
        & \includegraphics[scale=0.12, valign=t]{figures/supp/start_pick_place_end/o12-b2-s2-v0/color_000089.jpg}
        & \includegraphics[scale=0.12, valign=t]{figures/supp/start_pick_place_end/o12-b2-s2-v0/color_000166.jpg}\\

        \hline

    \end{tabular}
    }
    \vspace{2mm}
    \caption{Evaluation videos for 9 tasks. We show one example video for each of the 9 tasks from our evaluation video dataset. The 4 frames showing for each video are the start frame, pick frame, place frame, and end frame in the video. We get the pick and place frame indexes from the in-contact object trajectory, where the pick frame index is when the in-contact object is first detected, and the place frame index is when the in-contact object is last detected in the video.}
    \label{tab:eval_video_task}
\end{table*}

\begin{table*}
    \vspace{-5mm}
    \resizebox{\textwidth}{!}{
    \begin{tabular}{|p{2cm}|c|c|p{0.3cm}|c|c|}
        
        \hline
        Settings & Video 1 - Start Frame &  Video 1 - End Frame  && Video 2 - Start Frame  &  Video 2 - End Frame \\
        
        \hline
        Different view
        & \includegraphics[scale=0.12, valign=t]{figures/supp/start_pick_place_end/o0-b0-s1-v0/color_000000.jpg}
        & \includegraphics[scale=0.12, valign=t]{figures/supp/start_pick_place_end/o0-b0-s1-v0/color_000220.jpg}
        &
        & \includegraphics[scale=0.12, valign=t]{figures/supp/start_pick_place_end/o0-b0-s1-v1/color_000000.jpg}
        & \includegraphics[scale=0.12, valign=t]{figures/supp/start_pick_place_end/o0-b0-s1-v1/color_000176.jpg}
        \\

        \hline
        Different Background
        & \includegraphics[scale=0.12, valign=t]{figures/supp/start_pick_place_end/o0-b0-s1-v0/color_000000.jpg}
        & \includegraphics[scale=0.12, valign=t]{figures/supp/start_pick_place_end/o0-b0-s1-v0/color_000220.jpg}
        &
        & \includegraphics[scale=0.12, valign=t]{figures/supp/start_pick_place_end/o0-b2-s1-v0/color_000000.jpg}
        & \includegraphics[scale=0.12, valign=t]{figures/supp/start_pick_place_end/o0-b2-s1-v0/color_000176.jpg}\\

        \hline
        Different Slot Occlustion
        & \includegraphics[scale=0.12, valign=t]{figures/supp/start_pick_place_end/o0-b0-s1-v0/color_000000.jpg}
        & \includegraphics[scale=0.12, valign=t]{figures/supp/start_pick_place_end/o0-b0-s1-v0/color_000220.jpg}
        &
        & \includegraphics[scale=0.12, valign=t]{figures/supp/start_pick_place_end/o0-b0-s3-v0/color_000000.jpg}
        & \includegraphics[scale=0.12, valign=t]{figures/supp/start_pick_place_end/o0-b0-s3-v0/color_000171.jpg}
        \\

        \hline
        Settings & Video 1 - Start Frame &  Video 1 - End Frame  && Video 2 - Start Frame  &  Video 2 - End Frame \\
        
        \hline
        Different view
        & \includegraphics[scale=0.12, valign=t]{figures/supp/start_pick_place_end/o12-b0-s0-v0/color_000000.jpg}
        & \includegraphics[scale=0.12, valign=t]{figures/supp/start_pick_place_end/o12-b0-s0-v0/color_000161.jpg}
        &
        & \includegraphics[scale=0.12, valign=t]{figures/supp/start_pick_place_end/o12-b0-s0-v1/color_000000.jpg}
        & \includegraphics[scale=0.12, valign=t]{figures/supp/start_pick_place_end/o12-b0-s0-v1/color_000158.jpg}
        \\

        \hline
        Different Background
        & \includegraphics[scale=0.12, valign=t]{figures/supp/start_pick_place_end/o12-b0-s0-v0/color_000000.jpg}
        & \includegraphics[scale=0.12, valign=t]{figures/supp/start_pick_place_end/o12-b0-s0-v0/color_000161.jpg}
        &
        & \includegraphics[scale=0.12, valign=t]{figures/supp/start_pick_place_end/o12-b3-s0-v0/color_000000.jpg}
        & \includegraphics[scale=0.12, valign=t]{figures/supp/start_pick_place_end/o12-b3-s0-v0/color_000208.jpg}\\

        \hline
        Different Slot Occlustion
        & \includegraphics[scale=0.12, valign=t]{figures/supp/start_pick_place_end/o12-b0-s0-v0/color_000000.jpg}
        & \includegraphics[scale=0.12, valign=t]{figures/supp/start_pick_place_end/o12-b0-s0-v0/color_000161.jpg}
        &
        & \includegraphics[scale=0.12, valign=t]{figures/supp/start_pick_place_end/o12-b0-s2-v0/color_000000.jpg}
        & \includegraphics[scale=0.12, valign=t]{figures/supp/start_pick_place_end/o12-b0-s2-v0/color_000133.jpg}
        \\

        \hline

    \end{tabular}
    }
    \vspace{-2mm}
    \caption{Evaluation videos in 3 settings. 
    We show two example videos (``bread in toaster'', ``muffin in muffin tray'') in 3 different setting. For each task of our evaluation videos, we captured videos for 3 different settings to evaluate system performance when the human demonstration video and its paired robot-view video were in different conditions. 
    The 3 different settings we captured are (1) different views (two videos are captured from different camera views), (2) different backgrounds (two videos are captured from different backgrounds where we set up the two scenes on two different tables), and (3) different slot occlusions (two videos are captured from different slot occlusions, where one video with all other slots empty while the other video with some slots full).}
    \label{tab:eval_video_setting}
\end{table*}
\section{Additional Implementation Details}
\label{sec:implementation}

\subsection{Slot-Net}
We follow SAM's training recipe for finetuning the ViT-based encoder using a combination of Dice Loss and Cross-Entropy Loss. We train the model on an A100 GPU for 72K iterations. We use a learning rate of $1\times10^{-5}$ with a batch size of 8.

\subsection{CLIPort++}
We start with the released code from the authors. To prepare labels for training from our RGBD videos, we treat our videos as a one-step action with the corresponding task name as the language goal. And we treat the center of the object mask as the pick action location and the center of the slot mask as the place action location.
We use 80 videos for training and 64 videos for evaluation, ensuring there is no overlap in tasks, objects, and backgrounds. 
\section{Additional Results of \systemName}
\label{sec:slerp_results}



\subsection{Full Ablation Study Table}
Tab.~\ref{tab:vp_ablation} presents the full ablation study results evaluated all all the three different settings.
\begin{table*}[t]
  \resizebox{\textwidth}{!}{
  \centering
  
  \setlength{\tabcolsep}{4pt}
  \begin{tabular}{@{}l|ccc|ccccc|ccccc|ccccc@{}}
    \toprule
    \multirow{2}{*}{Method} & \multirow{2}{*}{SlotNet} & \multirow{2}{*}{SAM2} & \multirow{2}{*}{Mast3r}  
    & \multicolumn{5}{c}{Diff. View} &  \multicolumn{5}{c}{Diff. Background} & \multicolumn{5}{c}{Diff. Slot Occu.}  \\
    &&& 
    & Obj $\uparrow$ &  Slot $\uparrow$ & Prec. $\uparrow$ & CD $\downarrow$  & EMD $\downarrow$ 
    & Obj $\uparrow$ &  Slot $\uparrow$ & Prec. $\uparrow$ & CD $\downarrow$  & EMD $\downarrow$ 
    & Obj $\uparrow$ &  Slot $\uparrow$ & Prec. $\uparrow$ & CD $\downarrow$  & EMD $\downarrow$ \\
    \midrule
    Base design      & \xmark & \xmark & \xmark                           &28.53 &28.23  &24.72  &0.2289  &0.1183  &33.27  &23.40  &22.22  &0.1217  &0.0643  &27.71  &27.05  &26.47  &0.1499  &0.0809\\
    Ours w/o SlotNet  & \xmark & \cmark & \cmark                           &\textbf{73.85}  &29.63  &27.84  &0.0486  &0.0289  &\textbf{70.27}  &23.91  &18.80  &0.0630  &0.0367  &\textbf{68.12}  &26.33  &25.06  &0.0520  &0.0315\\
    Ours w/o SAM2    & \cmark & \xmark & \cmark         &31.05  &35.98  &23.75  &0.1219  &0.0667  &33.98  &30.87  &16.99  &0.0870  &0.0505  &28.62  &37.48  &22.97  &0.1012  &0.0575\\
    Ours w/o Mast3r   & \cmark & \cmark & \xmark &\textbf{73.85}  &\textbf{54.37}  &32.74  &0.0321  &0.0205  &\textbf{70.27}  &\textbf{44.70}  &\textbf{28.93}  &\textbf{0.0540}  &\textbf{0.0318}  &\textbf{68.12}  &\textbf{47.04}  &27.38  &0.0675  &0.0374\\
    Ours             & \cmark & \cmark         & \cmark                   &\textbf{73.85}  &\textbf{54.37}  &\textbf{36.40}  &\textbf{0.0282}  &\textbf{0.0182}  &\textbf{70.27}  &\textbf{44.70}  &25.39  &0.0573  &0.0323  &\textbf{68.12}  &\textbf{47.04}  &\textbf{30.30}  &\textbf{0.0334}  &\textbf{0.0223}\\
    
    \bottomrule
  \end{tabular}
  }
    \vspace{-2mm}
  \caption{
  \textbf{System Ablation Studies.} We conduct ablation studies on the key modules in \systemName. All metrics and test splits follow Table~\ref{tab:vp_comparison}. 
  Results show all the key modules help, and our full system performs the best on average across different settings.
  }
  \label{tab:vp_ablation}
      \vspace{-2mm}
\end{table*}

\subsection{Multi-slots analysis.}
Tab.~\ref{tab:diffslots} further provides 1-to-another slot placement results, placing into a different slot as human videos, and shows ours generalizes well.

\begin{table}[H]
\centering
  \footnotesize
  \renewcommand{\arraystretch}{0.3}
        \begin{tabular}{@{}l|ccccc}
        \toprule
        \multirow{2}{*}{} & \multicolumn{5}{c}{Diff Slot Placement} \\
        & Obj$\uparrow$ &  Slot$\uparrow$ & Prec.$\uparrow$ & CD$\downarrow$  & EMD$\downarrow$ \\
        \midrule
        Ours         &67.70  &42.62  &23.14  &0.0575 &0.0350 \\
        \bottomrule
      \end{tabular}
\hspace{-1.7em}
\caption{Object placement into a different slot as in the human video.}
\label{tab:diffslots}
\end{table}
\vspace{-1.4em}

To evaluate DINOv2+SAM 1-to-N slot re-identification, we annotated all N slots, and Table~\ref{tab:multislots} shows that our method does reliably well in finding all equivalent slots. 

\vspace{-1.0em}
\begin{table}[H]
\centering
  \footnotesize
  \renewcommand{\arraystretch}{0.3}
    \begin{tabular}{@{}l|cc|ccc|cccc@{}}
    \toprule
    &  \multicolumn{2}{c}{Diff View} &  &  \multicolumn{2}{c}{Diff Bg} &  & \multicolumn{2}{c}{Diff Slot Occ.} \\
    & mIoU        & AP             && mIoU      & AP          && mIoU     & AP  \\ \midrule
    Ours & 68.3 & 46.3 & & 66.8 & 44.2 & & 66.0 & 43.3 \\
    \bottomrule
    \end{tabular}
\hspace{-1.7em}
\caption{Multi-slot mask identification evaluation.}
\label{tab:multislots}
\end{table}
\vspace{-1.5em}

\subsection{More baseline comparison results}
We show more qualitative comparison results for baselines in Fig.~\ref{fig:supp_vp_baseline}. 
 \begin{figure*}[!tbp]
\centering
\includegraphics[width=\linewidth]{figures/supp/supp_vp_baseline.pdf}
\vspace{-6mm}
\caption{\textbf{Qualitative Comparison for baselines.} We compare our method to baselines and present side-by-side results on six examples. 
For each example, the top row displays 2D slot prediction results, while the bottom row shows 6-DoF relative pose predictions by projecting the object point cloud onto the slots.}
\label{fig:supp_vp_baseline}
\vspace{-3mm}
\end{figure*}

\subsection{More ablation comparison results}
We show more qualitative comparison results for ablations in Fig.~\ref{fig:supp_vp_ablation}. 
 \begin{figure*}[!tbp]
\centering
\includegraphics[width=\linewidth]{figures/supp/supp_vp_ablation.pdf}
\vspace{-6mm}
\caption{\textbf{Qualitative Comparison for ablations.} We compare our method to ablations and present side-by-side results on six examples. 
For each example, the top row displays 2D slot prediction results, while the bottom row shows 6-DoF relative pose predictions by projecting the object point cloud onto the slots.}
\label{fig:supp_vp_ablation}
\vspace{-3mm}
\end{figure*}

\subsection{Limitations}

\begin{table*}
    \vspace{-5mm}
    \resizebox{\textwidth}{!}{
    \begin{tabular}{|c|c|c|c|c|c|c|c|}
        
        \hline
        \huge Start image & \huge Image Diff & \huge  Obj-Box-as-Slot & \huge Obj GT mask & \huge SlotNet-endimg  &  \huge Ours & \huge GT   \\
        
        \hline
        \includegraphics[scale=0.12, valign=t]{figures/supp/slotnet_results/start/o0-b3-s2-v1.jpg}
        & \includegraphics[scale=0.12, valign=t]{figures/supp/slotnet_results/imagediff_thd50__seen_real_data/o0-b3-s2-v1.png}
        & \includegraphics[scale=0.12, valign=t]{figures/supp/slotnet_results/pickobj_bbox_asslot_mask__seen_real_data/o0-b3-s2-v1.png}
        & \includegraphics[scale=0.12, valign=t]{figures/supp/slotnet_results/pickobj_gt_mask__seen_real_data/o0-b3-s2-v1.png}
        & \includegraphics[scale=0.12, valign=t]{figures/supp/slotnet_results/slotnet_endimage__seen_real_data/o0-b3-s2-v1.png}
        & \includegraphics[scale=0.12, valign=t]{figures/supp/slotnet_results/slotnet_imgdiff__seen_real_data/o0-b3-s2-v1.png}
        & \includegraphics[scale=0.12, valign=t]{figures/supp/slotnet_results/gt/o0-b3-s2-v1_gt.png}
        \\

        \hline
        \includegraphics[scale=0.12, valign=t]{figures/supp/slotnet_results/start/o4-b1-s0-v0.jpg}
        & \includegraphics[scale=0.12, valign=t]{figures/supp/slotnet_results/imagediff_thd50__seen_real_data/o4-b1-s0-v0.png}
        & \includegraphics[scale=0.12, valign=t]{figures/supp/slotnet_results/pickobj_bbox_asslot_mask__seen_real_data/o4-b1-s0-v0.png}
        & \includegraphics[scale=0.12, valign=t]{figures/supp/slotnet_results/pickobj_gt_mask__seen_real_data/o4-b1-s0-v0.png}
        & \includegraphics[scale=0.12, valign=t]{figures/supp/slotnet_results/slotnet_endimage__seen_real_data/o4-b1-s0-v0.png}
        & \includegraphics[scale=0.12, valign=t]{figures/supp/slotnet_results/slotnet_imgdiff__seen_real_data/o4-b1-s0-v0.png}
        & \includegraphics[scale=0.12, valign=t]{figures/supp/slotnet_results/gt/o4-b1-s0-v0_gt.png}
        \\

        \hline
        \includegraphics[scale=0.12, valign=t]{figures/supp/slotnet_results/start/o8-b1-s2-v1.jpg}
        & \includegraphics[scale=0.12, valign=t]{figures/supp/slotnet_results/imagediff_thd50__seen_real_data/o8-b1-s2-v1.png}
        & \includegraphics[scale=0.12, valign=t]{figures/supp/slotnet_results/pickobj_bbox_asslot_mask__seen_real_data/o8-b1-s2-v1.png}
        & \includegraphics[scale=0.12, valign=t]{figures/supp/slotnet_results/pickobj_gt_mask__seen_real_data/o8-b1-s2-v1.png}
        & \includegraphics[scale=0.12, valign=t]{figures/supp/slotnet_results/slotnet_endimage__seen_real_data/o8-b1-s2-v1.png}
        & \includegraphics[scale=0.12, valign=t]{figures/supp/slotnet_results/slotnet_imgdiff__seen_real_data/o8-b1-s2-v1.png}
        & \includegraphics[scale=0.12, valign=t]{figures/supp/slotnet_results/gt/o8-b1-s2-v1_gt.png}
        \\

        \hline
        \includegraphics[scale=0.12, valign=t]{figures/supp/slotnet_results/start/o10-b1-s2-v1.jpg}
        & \includegraphics[scale=0.12, valign=t]{figures/supp/slotnet_results/imagediff_thd50__seen_real_data/o10-b1-s2-v1.png}
        & \includegraphics[scale=0.12, valign=t]{figures/supp/slotnet_results/pickobj_bbox_asslot_mask__seen_real_data/o10-b1-s2-v1.png}
        & \includegraphics[scale=0.12, valign=t]{figures/supp/slotnet_results/pickobj_gt_mask__seen_real_data/o10-b1-s2-v1.png}
        & \includegraphics[scale=0.12, valign=t]{figures/supp/slotnet_results/slotnet_endimage__seen_real_data/o10-b1-s2-v1.png}
        & \includegraphics[scale=0.12, valign=t]{figures/supp/slotnet_results/slotnet_imgdiff__seen_real_data/o10-b1-s2-v1.png}
        & \includegraphics[scale=0.12, valign=t]{figures/supp/slotnet_results/gt/o10-b1-s2-v1_gt.png}
        \\

        \hline
        \includegraphics[scale=0.12, valign=t]{figures/supp/slotnet_results/start/o11-b0-s3-v0.jpg}
        & \includegraphics[scale=0.12, valign=t]{figures/supp/slotnet_results/imagediff_thd50__seen_real_data/o11-b0-s3-v0.png}
        & \includegraphics[scale=0.12, valign=t]{figures/supp/slotnet_results/pickobj_bbox_asslot_mask__seen_real_data/o11-b0-s3-v0.png}
        & \includegraphics[scale=0.12, valign=t]{figures/supp/slotnet_results/pickobj_gt_mask__seen_real_data/o11-b0-s3-v0.png}
        & \includegraphics[scale=0.12, valign=t]{figures/supp/slotnet_results/slotnet_endimage__seen_real_data/o11-b0-s3-v0.png}
        & \includegraphics[scale=0.12, valign=t]{figures/supp/slotnet_results/slotnet_imgdiff__seen_real_data/o11-b0-s3-v0.png}
        & \includegraphics[scale=0.12, valign=t]{figures/supp/slotnet_results/gt/o11-b0-s3-v0_gt.png}
        \\

        \hline
        \includegraphics[scale=0.12, valign=t]{figures/supp/slotnet_results/start/o12-b1-s0-v1.jpg}
        & \includegraphics[scale=0.12, valign=t]{figures/supp/slotnet_results/imagediff_thd50__seen_real_data/o12-b1-s0-v1.png}
        & \includegraphics[scale=0.12, valign=t]{figures/supp/slotnet_results/pickobj_bbox_asslot_mask__seen_real_data/o12-b1-s0-v1.png}
        & \includegraphics[scale=0.12, valign=t]{figures/supp/slotnet_results/pickobj_gt_mask__seen_real_data/o12-b1-s0-v1.png}
        & \includegraphics[scale=0.12, valign=t]{figures/supp/slotnet_results/slotnet_endimage__seen_real_data/o12-b1-s0-v1.png}
        & \includegraphics[scale=0.12, valign=t]{figures/supp/slotnet_results/slotnet_imgdiff__seen_real_data/o12-b1-s0-v1.png}
        & \includegraphics[scale=0.12, valign=t]{figures/supp/slotnet_results/gt/o12-b1-s0-v1_gt.png}
        \\

        \hline        
    \end{tabular}
    }
    \vspace{-2mm}
    \caption{More comparison results for \modelName on seen tasks. We show more examples for comparison results between \modelName and other baselines or heuristics.}
    \label{tab:slotnet_results}
\end{table*}

\begin{table*}
    \vspace{-5mm}
    \resizebox{\textwidth}{!}{
    \begin{tabular}{|c|c|c|c|c|c|c|}
        
        \hline
        \huge Start image & \huge Ours(\modelName) & \huge  GT &  & \huge Start  &  \huge Ours(\modelName) & \huge GT   \\
        
        \hline
        \includegraphics[scale=0.12,   valign=t]{figures/supp/slotnet_results_random/start/o0-b0-s2-v0.jpg}
        & \includegraphics[scale=0.12, valign=t]{figures/supp/slotnet_results_random/slotnet_imgdiff__seen_real_data/o0-b0-s2-v0.png}
        & \includegraphics[scale=0.12, valign=t]{figures/supp/slotnet_results_random/gt/o0-b0-s2-v0_gt.png}
        &
        & \includegraphics[scale=0.12, valign=t]{figures/supp/slotnet_results_random/start/o0-b1-s0-v0.jpg}
        & \includegraphics[scale=0.12, valign=t]{figures/supp/slotnet_results_random/slotnet_imgdiff__seen_real_data/o0-b1-s0-v0.png}
        & \includegraphics[scale=0.12, valign=t]{figures/supp/slotnet_results_random/gt/o0-b1-s0-v0_gt.png}
        \\

        \hline
        \includegraphics[scale=0.12,   valign=t]{figures/supp/slotnet_results_random/start/o0-b2-s1-v1.jpg}
        & \includegraphics[scale=0.12, valign=t]{figures/supp/slotnet_results_random/slotnet_imgdiff__seen_real_data/o0-b2-s1-v1.png}
        & \includegraphics[scale=0.12, valign=t]{figures/supp/slotnet_results_random/gt/o0-b2-s1-v1_gt.png}
        &
        & \includegraphics[scale=0.12, valign=t]{figures/supp/slotnet_results_random/start/o4-b1-s3-v1.jpg}
        & \includegraphics[scale=0.12, valign=t]{figures/supp/slotnet_results_random/slotnet_imgdiff__seen_real_data/o4-b1-s3-v1.png}
        & \includegraphics[scale=0.12, valign=t]{figures/supp/slotnet_results_random/gt/o4-b1-s3-v1_gt.png}
        \\

        \hline
        \includegraphics[scale=0.12,   valign=t]{figures/supp/slotnet_results_random/start/o4-b3-s3-v0.jpg}
        & \includegraphics[scale=0.12, valign=t]{figures/supp/slotnet_results_random/slotnet_imgdiff__seen_real_data/o4-b3-s3-v0.png}
        & \includegraphics[scale=0.12, valign=t]{figures/supp/slotnet_results_random/gt/o4-b3-s3-v0_gt.png}
        &
        & \includegraphics[scale=0.12, valign=t]{figures/supp/slotnet_results_random/start/o4-b3-s3-v1.jpg}
        & \includegraphics[scale=0.12, valign=t]{figures/supp/slotnet_results_random/slotnet_imgdiff__seen_real_data/o4-b3-s3-v1.png}
        & \includegraphics[scale=0.12, valign=t]{figures/supp/slotnet_results_random/gt/o4-b3-s3-v1_gt.png}
        \\

        \hline
        \includegraphics[scale=0.12,   valign=t]{figures/supp/slotnet_results_random/start/o8-b0-s2-v1.jpg}
        & \includegraphics[scale=0.12, valign=t]{figures/supp/slotnet_results_random/slotnet_imgdiff__seen_real_data/o8-b0-s2-v1.png}
        & \includegraphics[scale=0.12, valign=t]{figures/supp/slotnet_results_random/gt/o8-b0-s2-v1_gt.png}
        &
        & \includegraphics[scale=0.12, valign=t]{figures/supp/slotnet_results_random/start/o8-b1-s2-v1.jpg}
        & \includegraphics[scale=0.12, valign=t]{figures/supp/slotnet_results_random/slotnet_imgdiff__seen_real_data/o8-b1-s2-v1.png}
        & \includegraphics[scale=0.12, valign=t]{figures/supp/slotnet_results_random/gt/o8-b1-s2-v1_gt.png}
        \\

        \hline
        \includegraphics[scale=0.12,   valign=t]{figures/supp/slotnet_results_random/start/o8-b2-s1-v0.jpg}
        & \includegraphics[scale=0.12, valign=t]{figures/supp/slotnet_results_random/slotnet_imgdiff__seen_real_data/o8-b2-s1-v0.png}
        & \includegraphics[scale=0.12, valign=t]{figures/supp/slotnet_results_random/gt/o8-b2-s1-v0_gt.png}
        &
        & \includegraphics[scale=0.12, valign=t]{figures/supp/slotnet_results_random/start/o10-b2-s0-v1.jpg}
        & \includegraphics[scale=0.12, valign=t]{figures/supp/slotnet_results_random/slotnet_imgdiff__seen_real_data/o10-b2-s0-v1.png}
        & \includegraphics[scale=0.12, valign=t]{figures/supp/slotnet_results_random/gt/o10-b2-s0-v1_gt.png}
        \\

        \hline
        \includegraphics[scale=0.12,   valign=t]{figures/supp/slotnet_results_random/start/o10-b2-s3-v0.jpg}
        & \includegraphics[scale=0.12, valign=t]{figures/supp/slotnet_results_random/slotnet_imgdiff__seen_real_data/o10-b2-s3-v0.png}
        & \includegraphics[scale=0.12, valign=t]{figures/supp/slotnet_results_random/gt/o10-b2-s3-v0_gt.png}
        &
        & \includegraphics[scale=0.12, valign=t]{figures/supp/slotnet_results_random/start/o10-b3-s3-v1.jpg}
        & \includegraphics[scale=0.12, valign=t]{figures/supp/slotnet_results_random/slotnet_imgdiff__seen_real_data/o10-b3-s3-v1.png}
        & \includegraphics[scale=0.12, valign=t]{figures/supp/slotnet_results_random/gt/o10-b3-s3-v1_gt.png}
        \\

        \hline
        \includegraphics[scale=0.12,   valign=t]{figures/supp/slotnet_results_random/start/o11-b2-s1-v1.jpg}
        & \includegraphics[scale=0.12, valign=t]{figures/supp/slotnet_results_random/slotnet_imgdiff__seen_real_data/o11-b2-s1-v1.png}
        & \includegraphics[scale=0.12, valign=t]{figures/supp/slotnet_results_random/gt/o11-b2-s1-v1_gt.png}
        &
        & \includegraphics[scale=0.12, valign=t]{figures/supp/slotnet_results_random/start/o12-b0-s0-v0.jpg}
        & \includegraphics[scale=0.12, valign=t]{figures/supp/slotnet_results_random/slotnet_imgdiff__seen_real_data/o12-b0-s0-v0.png}
        & \includegraphics[scale=0.12, valign=t]{figures/supp/slotnet_results_random/gt/o12-b0-s0-v0_gt.png}
        \\

        \hline
        \includegraphics[scale=0.12,   valign=t]{figures/supp/slotnet_results_random/start/o12-b0-s1-v0.jpg}
        & \includegraphics[scale=0.12, valign=t]{figures/supp/slotnet_results_random/slotnet_imgdiff__seen_real_data/o12-b0-s1-v0.png}
        & \includegraphics[scale=0.12, valign=t]{figures/supp/slotnet_results_random/gt/o12-b0-s1-v0_gt.png}
        &
        & \includegraphics[scale=0.12, valign=t]{figures/supp/slotnet_results_random/start/o12-b3-s2-v1.jpg}
        & \includegraphics[scale=0.12, valign=t]{figures/supp/slotnet_results_random/slotnet_imgdiff__seen_real_data/o12-b3-s2-v1.png}
        & \includegraphics[scale=0.12, valign=t]{figures/supp/slotnet_results_random/gt/o12-b3-s2-v1_gt.png}
        \\

        \hline        
    \end{tabular}
    }
    \vspace{-2mm}
    \caption{Random results for \modelName on seen tasks. We show randomly sampled results from \modelName on seen tasks.}
    \label{tab:slotnet_results_random}
\end{table*}
Our system, \systemName, demonstrates superior performance compared to baseline methods. However, there are areas that require improvement. Firstly, \systemName is modular, with each module relying on the output of the preceding one, which may result in compounding errors. If a preceding module fails to provide accurate outputs, subsequent results may be adversely affected. Secondly, while the modules currently employed are not without flaws, they can be easily replaced with more advanced methods.

For instance, \modelName necessitates that the \textit{start} and \textit{end} images have minimal changes aside from the pick object; future methodologies could relax this requirement, enabling slot detection in more general settings. Additionally, SAM2 is a tracking method repurposed for re-identification, and methods specifically designed for re-identification would likely yield better results. Lastly, while Mast3r is effective for object-level matching, it struggles with slot-level matching, indicating that improved slot-level matching algorithms could enhance accuracy.

\section{Additional Results of \modelName}
\label{sec:slotnet_results}

\subsection{Additional results of \modelName}
We provide more \modelName comparison results in Tab.~\ref{tab:slotnet_results} and random results in Tab.~\ref{tab:slotnet_results_random}.

\subsection{Limitations}

\modelName has limitations in slot segmentation across varying conditions due to its training data constraints:
(1) It requires that the input (\textit{start}, \textit{end}) images have minimal changes in viewpoint.
(2) It necessitates that the pick object is the sole changing element between (\textit{start}, \textit{end}) images, and neither humans nor human hands are present.
\modelName may exhibit the following issues: (1) While it can generalize to unseen tasks, its performance may degrade. (2) The predicted slot may encompass multiple slots or extend beyond the ground truth. (3) It may incorrectly identify pick objects as slots.

{
    \small
    \bibliographystyle{ieeenat_fullname}
    \bibliography{main}
}
